\documentclass[lettersize,journal]{IEEEtran}
\usepackage{amsmath,amsfonts}
\usepackage{algorithm}
\usepackage{array}
\usepackage[caption=false,font=normalsize,labelfont=sf,textfont=sf]{subfig}
\usepackage{textcomp}
\usepackage{stfloats}
\usepackage{url}
\usepackage{verbatim}
\usepackage{graphicx}
\usepackage{cite}
\usepackage{algpseudocode}
\usepackage{enumitem}
\usepackage{hyperref}
\usepackage{amssymb}
\usepackage{colortbl}  
\usepackage{xcolor}  
\usepackage{makecell}  
\usepackage{multirow}  
\usepackage{pifont}

\begin{document}
\title{VLBiMan++: Expanding the Generalization Boundary of Vision-Language Anchored One-Shot Bimanual Manipulation}

\author{Huayi Zhou,~\IEEEmembership{Member,~IEEE,} Wei Gao, Yiyang Han, Kui Jia,~\IEEEmembership{Member,~IEEE,} Hui Huang,~\IEEEmembership{Senior Member,~IEEE} 
\thanks{H. Zhou and H. Huang are with Guangdong Provincial Key Laboratory of Visual Media and Multidimensional Intelligence, College of Computer Science and Software Engineering, Shenzhen University. K. Jia is with School of Data Science, The Chinese University of Hong Kong, Shenzhen. W. Gao and K. Jia are with DexForce, Shenzhen. Y. Han is with the Faculty of Engineering, Department of Computing, Imperial College London.}
\thanks{This work was supported by the Guangdong Provincial Key Field R\&D Program (Project No. 20240104), and was also funded by the Shenzhen Science and Technology Major Project (No. 202402002 and ZDCT20250901113000001).}
\thanks{Manuscript received April 25, 2026; revised August 25, 2026.}}

\markboth{Journal of \LaTeX\ Class Files,~Vol.~14, No.~8, August~2026}%
{Shell \MakeLowercase{\textit{et al.}}: A Sample Article Using IEEEtran.cls for IEEE Journals}

\IEEEpubid{0000--0000/00\$00.00~\copyright~2026 IEEE}


\maketitle
\begin{abstract}
Generalizable bimanual robotic manipulation requires a reusable task prior that can persist across increasingly diverse tasks, objects, scenes, embodiments, and execution conditions, thus avoiding the prohibitive cost of large-scale teleoperated demonstrations and policy retraining. In this work, we present VLBiMan++, an extended framework that expands the generalization boundary of vision-language anchored one-shot bimanual manipulation. Starting from a single human demonstration, VLBiMan++ performs task-aware decomposition to identify reusable and adaptable skill components, and employs vision-language grounded geometric adaptation to transfer these skills to novel configurations without retraining. Building on this foundation, we systematically extend generalization along five dimensions: task generalization through diverse and long-horizon skill compositions; object generalization across unseen categories, varying geometries, and more complex articulated or deformable objects; scene generalization under clutter, occlusion, and dynamic interference; embodiment generalization across heterogeneous dual-arm robotic platforms; and deployment generalization through prolonged closed-loop execution under repeated external perturbations. To support this broader scope, we further introduce object-state-aware adaptation and lightweight trajectory optimization mechanisms that accommodate changes beyond simple rigid-body 6-DoF pose variations while preserving reliable bimanual coordination. Extensive real-world experiments demonstrate that VLBiMan++ maintains strong task success and adaptation capability across these increasingly challenging settings. Overall, VLBiMan++ advances one-shot bimanual manipulation from demonstrating isolated transferability toward a more systematic and scalable framework for generalization across tasks, objects, scenes, embodiments, and long-term deployment conditions. The project link is \url{https://hnuzhy.github.io/projects/VLBiManPlus}.
\end{abstract}

\section{Introduction} 

\IEEEPARstart{R}{ecent} years have witnessed remarkable progress in embodied robotic manipulation, particularly through visuomotor imitation learning based on large-scale teleoperated demonstrations \cite{fang2024rh20t, khazatsky2024droid, o2024open, bu2025agibot}. By collecting large quantities of real-world demonstrations, recent Vision-Language-Action (VLA) models \cite{team2024octo, kim2024openvla, lin2025data} learn to directly map multimodal observations to robot actions, implicitly encoding task-, object-, and environment-specific complexities into high-dimensional latent representations. This paradigm has demonstrated impressive capabilities on challenging high-degree-of-freedom manipulation problems, including bimanual and long-horizon tasks, as evidenced by the ALOHA family \cite{zhao2023learning, fu2024mobile, aldaco2024aloha, zhao2024aloha}, RDT-1B \cite{liu2025rdt}, $\pi_0$ \cite{black2025pi0}, and FAST \cite{pertsch2025fast}. However, its scalability remains fundamentally constrained by the cost of data collection and policy retraining: adapting an existing policy to new tasks, object configurations, or robotic embodiments often requires substantial demonstration re-collection and retraining. Such a paradigm is therefore difficult to scale to the unstructured combinations of tasks, objects, scenes, and robot platforms encountered in real-world deployment.

An alternative direction is to exploit pretrained foundation models and structured intermediate representations to decouple semantic understanding from low-level control. Recent modular robotic systems leverage Large Language Models (LLMs) \cite{achiam2023gpt} and Vision-Language Models (VLMs) \cite{radford2021learning, xiao2024florence} for task interpretation, semantic grounding, and object-centric perception, while delegating motion generation to rule-based controllers, learned visuomotor policies, or reusable skill libraries \cite{chi2023diffusion, ze2024dp3, yang2024equibot}. Besides, reinforcement learning in simulation also serves as a strategy for learning skill-specific controllers \cite{xie2020deep, chen2022towards, yuan2024learning}. Another particularly effective strategy is to introduce compact and transferable geometric representations, such as keypoints, affordances, and object-centric correspondences, as a bridge between perception and action. For example, ReKep \cite{huang2024rekep} constrains trajectory optimization using predicted relation points, MOKA \cite{fang2024moka} identifies functional regions through multimodal reasoning, and RobotPoint \cite{yuan2024robopoint} extracts task-relevant point clusters from visual observations. These studies demonstrate that object-centric abstractions can substantially improve manipulation transferability across changing viewpoints and instances. Nevertheless, existing approaches typically focus on a limited aspect of generalization and often depend on either extensive task-specific data, specialized perception models, or planning pipelines whose execution reliability can deteriorate as the manipulation scenario becomes increasingly diverse.

\IEEEpubidadjcol

\begin{figure*}
	\begin{center}
           \includegraphics[width=1.0\linewidth]{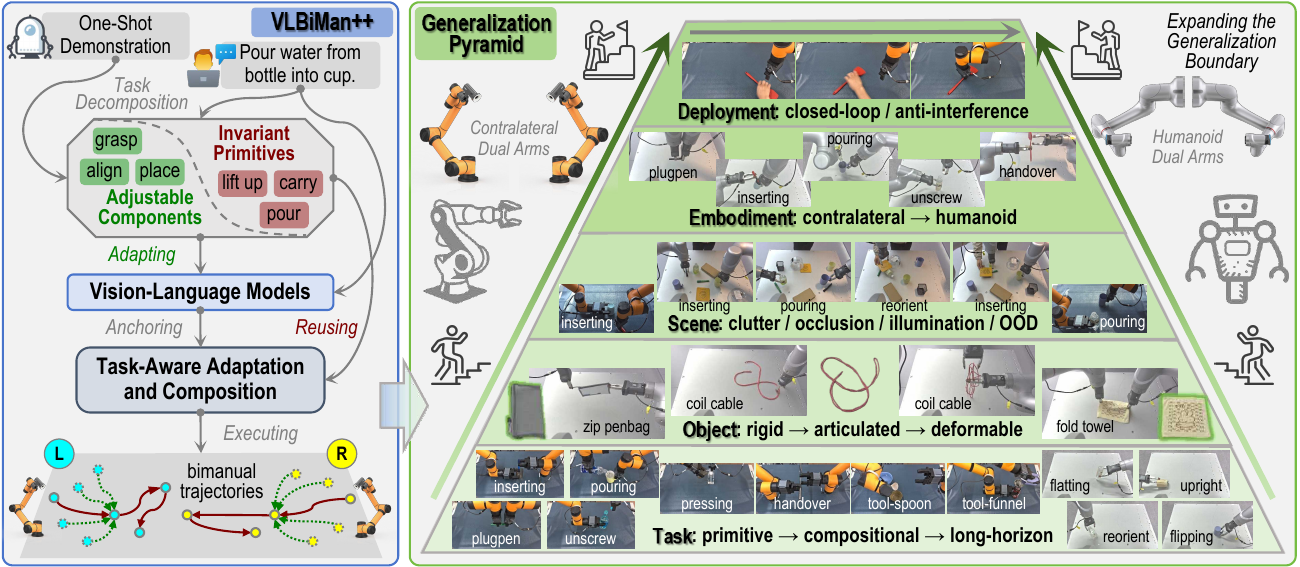}
	\vspace{-20pt}
	\caption{\textit{Left}: Taking pouring water as an example, we sketch the entire process of VLBiMan++ based on the one-shot demonstration. \textit{Right}: Diagram illustration of the proposed \textbf{Generalization Pyramid}: \textbf{Task} $\rightarrow$ \textbf{Object} $\rightarrow$ \textbf{Scene} $\rightarrow$ \textbf{Embodiment} $\rightarrow$ \textbf{Deployment}. VLBiMan++ aims to investigate how far a single human demonstration can serve as a reusable manipulation prior across increasingly diverse tasks, objects, scenes, embodiments, and execution conditions.}
           \label{teaser}
	\vspace{-15pt}
	\end{center}
\end{figure*}

Our previous conference work, \textbf{VLBiMan} \cite{zhou2026vlbiman}, explored a complementary hypothesis: rather than repeatedly relearning or replanning an entire manipulation policy, a single human demonstration can serve as a structured prior from which the essential and reusable parts of a task are extracted. VLBiMan performs (1) \textit{Task-Aware Bimanual Decomposition}, (2) \textit{Vision-Language Anchored Adaptation}, and (3) \textit{Autonomous Trajectory Composition} to separate invariant skill components from scene-dependent variations, align the latter through VLM-grounded object representations, and synthesize executable trajectories without policy retraining. The central design principle is that \textit{\textbf{what to achieve matters more than how to execute it}}: task-relevant spatial and functional relationships should be preserved, whereas incidental absolute poses and trajectories can be adapted. This formulation enables effective transfer from a single demonstration across object placements, category-level instances, long-horizon skill compositions, and heterogeneous robotic embodiments. However, the original VLBiMan \cite{zhou2026vlbiman} primarily establishes the feasibility of this paradigm on a controlled collection of rigid-object manipulation tasks. A more fundamental question therefore remains: \textit{how far can a one-shot human demonstration generalize when the task, object, scene, embodiment, and execution conditions progressively depart from those observed in the demonstration?}

In this journal extension, we answer above-mentioned question by introducing \textbf{VLBiMan++}, which expands the generalization boundary of vision-language anchored one-shot bimanual robotic manipulation along a systematic \textbf{Generalization Pyramid}: \textit{Task Generalization} $\rightarrow$ \textit{Object Generalization} $\rightarrow$ \textit{Scene Generalization} $\rightarrow$ \textit{Embodiment Generalization} $\rightarrow$ \textit{Deployment Generalization}. At the \textit{\textbf{task}} level, we substantially broaden the experimental suite to cover more diverse bimanual behaviors, multi-stage interactions, and long-horizon skill compositions, probing whether atomic skills extracted from a single demonstration can support broader task structures. At the \textit{\textbf{object}} level, we move beyond seen and category-level novel rigid objects to more challenging unseen object types and investigate articulated and deformable objects, extending the original adaptation formulation toward changes in object configuration and physical state rather than simple rigid-body pose variation. At the \textit{\textbf{scene}} level, we evaluate VLBiMan++ under cluttered environments containing semantic distractors, partial occlusions, and spatial interference, testing whether vision-language anchoring can continuously identify and track the task-relevant objects despite substantial perceptual ambiguity. At the \textit{\textbf{embodiment}} level, we further validate transfer across heterogeneous dual-arm platforms, demonstrating that the task prior extracted from a human demonstration can remain effective under substantially different kinematic and hardware configurations. Finally, at the deployment level, we conduct prolonged closed-loop execution with repeated object perturbations and external disturbances, examining not only instantaneous task success but also the stability and persistence of adaptation over extended operation.

To sum up, VLBiMan++ further extends the original framework to support broader object-state variations and deployment conditions while preserving its training-free, task-grounded philosophy. Beyond adapting rigid-body position and orientation, the enhanced object-centric adaptation accommodates articulated configurations and deformation, while the modular \textit{perception–adaptation–composition} pipeline enables repeated re-observation and re-alignment under object displacement and environmental disturbances. Accordingly, VLBiMan++ treats a human demonstration as a reusable task prior and systematically expands its generalization boundary from task composition and novel object states to cluttered scenes, heterogeneous embodiments, and prolonged closed-loop execution, providing a more comprehensive pathway toward scalable bimanual manipulation that reuses rather than repeatedly relearns human demonstrations. Our contributions are as follows:
\begin{itemize}[leftmargin=1em]
\item We introduce VLBiMan++, a substantial extension of our previous work VLBiMan, and formulate a Generalization Pyramid for one-shot bimanual manipulation that systematically studies generalization across tasks, objects, scenes, embodiments, and long-term deployment conditions.
\item We extend vision-language anchored adaptation beyond rigid-object pose transfer toward more complex object-state variations, including articulated and deformable manipulation, enabling the reusable prior to remain effective under substantially broader geometric and physical changes.
\item We establish a substantially expanded real-world evaluation suite covering unseen object categories, cluttered and dynamically perturbed scenes, new heterogeneous dual-arm embodiments, and prolonged closed-loop execution, providing comprehensive evidence of VLBiMan++'s robustness, versatility, and deployment stability.
\item We provide extensive analyses of the expanded generalization boundary, including failure modes, robustness under persistent disturbances, and cross-condition transferability, revealing both the practical strengths and remaining limitations of one-shot bimanual manipulation.
\end{itemize}



\section{Related Works}

\textbf{Generalizable Representations for Manipulation.}
Traditional robotic manipulation often relied on structured representations built upon strong priors \cite{kaelbling2013integrated, dantam2018incremental, migimatsu2020object, tyree20226}, such as object geometry or rigid-body assumptions, typically through 6D pose estimation or manually specified grasp configurations, which are difficult to scale to unstructured environments. With the rise of data-driven techniques, more flexible representations have emerged, including keypoints \cite{papagiannis2025rx, gao2024bi, wen2024any, grannen2021untangling}, affordances \cite{ju2024robo, nasiriany2025rt, zhao2023dualafford, tang2025uad}, dynamic flow fields \cite{colome2018dimensionality, weng2022fabricflownet}, and invariant object-centric correspondences \cite{ko2024learning, zhang2024one, zhang2023universal}. More recent works further extend these representations to articulated or deformable objects and leverage human demonstrations to retarget 3D hand trajectories to robots \cite{chen2024object, li2024okami, kerr2024robot, chen2025vividex}. However, such approaches often require private datasets, retraining, object-specific models, or complex retargeting pipelines, limiting their scalability across diverse object states. In contrast, VLBiMan++ employs lightweight object-centric representing points together with vision-language grounding and extends adaptation beyond rigid pose changes toward more diverse geometric, articulated, and deformable object variations, without retraining.

\textbf{Efficient Bimanual Robotic Manipulation.}
Recent advances in bimanual manipulation have demonstrated the capability of large Vision-Language-Action (VLA) models \cite{black2025pi0, liu2025rdt, pertsch2025fast} trained on extensive teleoperated demonstrations \cite{fang2024rh20t, khazatsky2024droid, o2024open, bu2025agibot}. However, scaling these systems to new tasks, objects, or embodiments often requires substantial additional data collection and retraining. Alternative efforts exploit large-scale Internet \cite{ponimatkin20256d, ye2025video2policy, bharadhwaj2024track2act} or egocentric human-hand videos \cite{zhan2024oakink2, liu2024taco, grauman2024ego, zhao2025taste, kareer2025egomimic}, yet the human-robot embodiment gap limits direct transfer. Other methods learn visuomotor policies \cite{chi2023diffusion, ze2024dp3} from relatively small real-world datasets, but their generalization remains constrained by task-specific training. One-shot imitation learning \cite{wen2022you, bahety2024screwmimic, zhou2025you, wang2025one, mao2023learning, liu2025one, biza2023one, zhou2026one, zhou2026yoto++} substantially reduces data requirements, while the high-dimensional and coordinated nature of bimanual manipulation remains challenging. In contrast, VLBiMan++ reuses task-invariant atomic skills extracted from a single bimanual demonstration and adaptively composes them across tasks, object states, scenes, and embodiments, providing a training-free route toward broader and more scalable generalization.

\textbf{Large Foundation Models for Robotics.}
Integrating LLMs and VLMs into robotics has become a prominent direction for building generalizable embodied agents \cite{ma2025vision, huang2025roboground, fangsam2act, feng2025reflective}. LLMs are commonly used for high-level task understanding and planning, such as decomposing instructions into executable subtasks or generating programs \cite{liang2023code, singh2023progprompt, szot2024large, huang2024copa}, while VLMs provide semantically grounded object detection and segmentation, often complemented by Visual Foundation Models (VFMs) \cite{oquab2024dinov2, ravi2025sam} for keypoints \cite{papagiannis2025rx, gao2024bi, wen2024any} and dense correspondences \cite{ko2024learning, zhang2024one}. Recent systems such as ReKep \cite{huang2024rekep}, MOKA \cite{fang2024moka}, RobotPoint \cite{yuan2024robopoint}, and RAM \cite{kuang2024ram} combine foundation models within modular \textit{perceive-understand-plan-act} pipelines for zero-shot manipulation. Despite their strong generalization, these methods often rely on extensive prompt engineering and ambiguous intermediate representations that require additional post-processing or planning. VLBiMan++ instead grounds manipulation adaptation in a precisely labeled one-shot demonstration, using VLMs to identify task-relevant objects and lightweight geometric anchors to selectively adapt only the components affected by object, scene, or embodiment changes. This design further extends foundation-model-guided manipulation toward articulated and deformable objects, cluttered environments, and prolonged closed-loop deployment without policy retraining.

Several concurrent studies have recently explored learning manipulation skills from a single human demonstration, sharing VLBiMan++'s goal of reducing demonstration and retraining costs. Multi-Task Trajectory Transfer (MT3) \cite{dreczkowski2025learning} decomposes trajectories into sequential alignment and interaction phases and employs retrieval-based generalization, enabling efficient transfer across a large collection of tasks. Behavior Prompting Policy (BPP) \cite{patel2026behavior} formulates demonstrations as behavior prompts and learns an in-context visuomotor policy that maps a demonstration and current observation directly to robot actions, while HOST \cite{chen2026robots} acquires skills from a single human video through self-grounded prediction and a shared task-progress manifold for embodiment adaptation. While these works demonstrate the strong potential of one-shot skill acquisition, VLBiMan++ differs in its emphasis on training-free, explicitly task-aware decomposition and vision-language anchored adaptation for bimanual manipulation, where invariant skill components are selectively reused and variable object-state components are geometrically adapted rather than inferred by a learned visuomotor policy. 



\section{Methodology}
\begin{figure*}
	\begin{center}
           \includegraphics[width=1.0\linewidth]{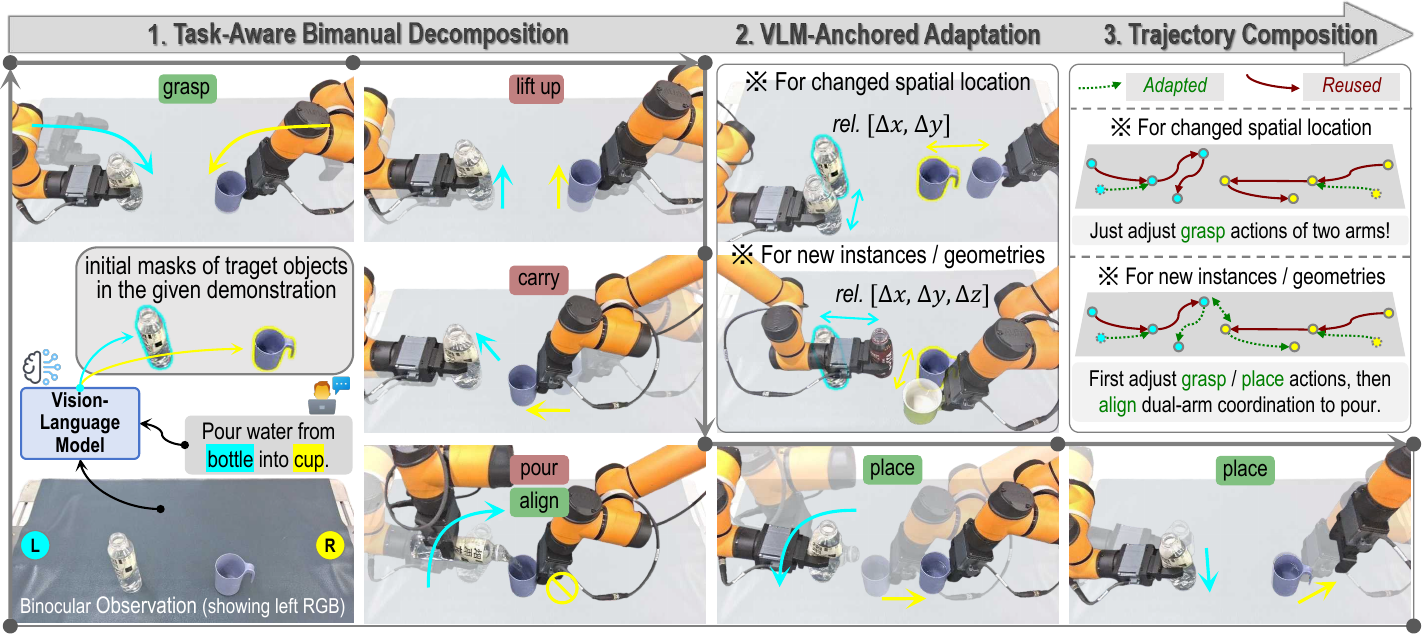}
	\vspace{-20pt}
	\caption{Framework of the expanded \textbf{V}ision-\textbf{L}anguage Anchored \textbf{Bi}manual \textbf{Man}ipulation (\textbf{VLBiMan++}). Taking the pouring water as an example, the paradigm consists of three stages (\textit{e.g.}, decomposition, adaptation, and composition) based on a given demonstration. VLBiMan++ can achieve generalization of unseen spatial placements, category-level new instances, non-rigid object state variations, and heterogeneous novel embodiments under the same task.}
           \label{framework}
	\vspace{-15pt}
	\end{center}
\end{figure*}

This section presents the complete pipeline of \textbf{VLBiMan++} (refer Fig.~\ref{framework}), extending vision-language anchored one-shot manipulation toward broader generalization across tasks, object states, scenes, and long-term deployment. We first formalize the problem and generalized manipulation states in Sec.~\ref{preliminaries}, and then introduce three interconnected components: (1) \textit{Task-Aware Bimanual Decomposition} in Sec.~\ref{TBD}, which decomposes a single demonstration into reusable atomic skills and characterizes their state dependence; (2) \textit{Vision-Language Anchored Adaptation} in Sec.~\ref{VLA}, which grounds task-relevant objects and adapts skills to rigid, articulated, and deformable object states; and (3) \textit{Autonomous Trajectory Composition} in Sec.~\ref{ATC}, which synthesizes executable bimanual trajectories through kinematic refinement, collision compensation, and closed-loop re-observation.

\subsection{Preliminaries}\label{preliminaries}
Given a concise textual description of a bimanual manipulation task and a single human demonstration collected in a canonical scene, our goal is to synthesize executable dual-arm trajectories for unseen conditions without policy retraining. Let $\mathcal{T}$ denote the task description and $\mathcal{D}=\{(\mathcal{O}_t,\mathcal{A}_t)\}_{t=1}^{T}$ denote the one-shot demonstration, where $\mathcal{O}_t$ is the multimodal observation at timestep $t$ and $\mathcal{A}_t$ is the corresponding bimanual action (\textit{e.g.}, 6-DoF end-effector poses of both arms, and gripper states). Given a novel scene state $\mathcal{S}_{\texttt{new}}$, which may involve changes in object identity, geometry, configuration, spatial arrangement, scene clutter, or robot embodiment, VLBiMan++ generates an adapted trajectory:
\begin{equation}
	\mathcal{F}_{\texttt{VLBiMan++}}: (\mathcal{T}, \mathcal{D}, \mathcal{S}_{\texttt{new}}) \mapsto \widetilde{\mathcal{D}}_{\texttt{new}} = \{\widetilde{\mathcal{A}}^{\texttt{new}}_t\}_{t=1}^{T'},
	\label{mapping}
\end{equation}
Here, $\mathcal{S}_{\texttt{new}}$ is not restricted to simple object relocation, scene rearrangement or category-level instance variation, but may further involve unseen object categories, articulated configurations, deformable states, cluttered environments, and changes in the underlying robot embodiment. The $\widetilde{\mathcal{A}}^\texttt{new}_t$ denotes the synthesized bimanual trajectory adapted to $\mathcal{S}_\texttt{new}$. Rather than reproducing the demonstration indiscriminately, VLBiMan++ preserves task-relevant invariant structure and selectively adapts state-dependent components. This requires addressing three core challenges: (1) \textit{Task-object semantic grounding}, which associates $\mathcal{T}$ with relevant objects ${o_k}_{k=1}^{K}$ through visual-language grounding; (2) \textit{Executable module decomposition}, which partitions $\mathcal{D}$ into temporally ordered motion primitives ${\mathcal{M}_i}$ with discrete boundaries ${t_i}$ and identifies their state dependence; and (3) \textit{Kinematically feasible trajectory composition}, which synthesizes adapted trajectories ${\widetilde{\mathcal{A}}^{\text{new}}_t}$ under geometric and physical constraints.

Accordingly, VLBiMan++ views a single given demonstration as a reusable task prior whose generalization spans progressively broader variations, from \textit{task} composition and \textit{object} state changes to \textit{scene} conditions, robot \textit{embodiment}, and long-term \textit{deployment}. Its objective is to maximize reuse of demonstration-derived manipulation knowledge while adapting only the components affected by the current scene state $\mathcal{S}_{\texttt{new}}$, thereby minimizing additional supervision, retraining, and full-trajectory replanning.

\subsection{Task-Aware Bimanual Decomposition}\label{TBD}
To construct reusable manipulation priors from a single demonstration, VLBiMan++ decomposes the demonstrated bimanual trajectory into temporally coherent segments and characterizes their dependence on object states. The procedure consists of \textit{Spatiotemporal Trajectory Segmentation}, \textit{Atomic Skill Extraction}, and \textit{State-Aware Skill Abstraction}.

\subsubsection{Spatiotemporal Trajectory Segmentation}
We record each one-shot demonstration using a third-person stereo RGB camera at 10 FPS, synchronously collecting the 6-DoF end-effector poses and gripper states of both arms. The resulting sequence is $\mathcal{D}\!=\!\{(\mathcal{O}_t,\mathcal{A}_t)\}_{t=1}^{T}$, where $\mathcal{A}_t\in\mathbb{R}^{14}$ contains the two end-effector poses and binary gripper states. We adopt a keypose-driven segmentation strategy inspired by discrete motion representations \cite{james2022coarse, shridhar2023perceiver, ma2024hierarchical, ke20243d}. Candidate keyposes are automatically detected from motion and state changes, including velocity or acceleration discontinuities and gripper transitions. Each keypose $\mathbf{w}_i$ defines a temporal slot $\tau_i=[t_i,t_{i+1}]$ and a candidate segment $\mathcal{M}_i=\{\mathcal{A}_t\}_{t\in\tau_i}$. Furthermore, because automatically detected keyposes may not always yield reliable waypoints for subsequent execution connected via inverse kinematics (IK) \cite{chitta2012moveit, schulman2014motion}, we apply lightweight human-in-the-loop refinement to adjust, insert, or remove a small number of waypoints when necessary. The refined segmentation is validated by real-robot rollout to ensure temporal consistency, spatial smoothness, and safe bimanual execution:
\begin{equation}
	\pi_{\texttt{seg}}:\mathcal{D} \rightarrow \{\mathcal{M}_i\}_{i=1}^{N},
	\label{segmentation}
\end{equation}
where $N$ denotes the number of temporally ordered motion segments, ranging from a few to over a dozen.

\subsubsection{Atomic Skill Extraction}
We next assign semantic roles to the segments according to the coupling between manipulated objects and robot end-effectors. Before grasping, the object and end-effector are spatially independent, making the corresponding motion generally sensitive to the current object state. After stable grasping, they form a coupled composite system, and subsequent motions can often be reused as task-level skills. Let $\texttt{bind}(o,r,t)\in\{0,1\}$ indicate whether object $o$ is coupled with end-effector $r$ at time $t$. Following the original decomposition in VLBiMan \cite{zhou2026vlbiman}, a segment $\mathcal{M}_i$ is regarded as invariant when the object remains coupled and its geometry is sufficiently consistent with the demonstration:
\begin{equation}
	\begin{aligned}
	&\forall t \in \tau_i, \; \texttt{bind}(o_k, r, t)=1, \text{and} \\
	& \quad\quad\quad\;\; \texttt{geometry}(o_k) \approx \texttt{geometry}(o_k^\texttt{demo}),
	\label{keypose}
	\end{aligned}
\end{equation}
where $\epsilon_g$ denotes the corresponding tolerance threshold for geometrically equivalent dimensions. Otherwise, the segment is considered potentially adaptable. The given demonstration is therefore decomposed into:
\begin{equation}
	\centering
	\mathcal{D} \Rightarrow \{\mathcal{M}_i^\texttt{inv}\}^{N_\texttt{inv}}_{i=1} \cup \{\mathcal{M}_j^\texttt{var}\}^{N_\texttt{var}}_{j=1}.
	\label{decompose}
\end{equation}
which are subsequently reused or adapted for novel scenes, tasks and object variations. Some illustrations on the pouring water task can be found in Fig.~\ref{framework} left.

\subsubsection{State-Aware Skill Abstraction}
VLBiMan++ further generalizes this decomposition by explicitly modeling whether a skill depends on the task-relevant object state $s_o$. Here, $s_o$ may describe pose and geometry for rigid objects, configuration for articulated objects, or state for deformable objects. We distinguish \textit{state-invariant} skills, whose functional execution remains valid despite changes in $s_o$, from \textit{state-conditioned} skills, whose execution must be adapted accordingly:
\begin{equation}
	\left\{ \begin{array}{ll}
	\mathcal{M}_i^{\texttt{SI}}\in\mathcal{S}^{\texttt{SI}}, &{\partial\mathcal{M}_i}/{\partial s_o}\approx 0, \\
	\mathcal{M}_j^{\texttt{SC}}(s_o)\in\mathcal{S}^{\texttt{SC}}, & {\partial\mathcal{M}_j}/{\partial s_o}\neq 0,
	\end{array}\right.
	\label{state_invariant_conditioned}
\end{equation}
Here, \textit{state-invariant} skills typically correspond to motions such as lifting, carrying, or transporting an already grasped object, whereas \textit{state-conditioned} skills include motions whose feasibility depends on object position, orientation, geometry, articulation, or deformation. Thus, rigid, articulated, and deformable objects can be handled within a unified abstraction: invariant structure is preserved as a reusable skill, while state-dependent components are exposed for subsequent adaptation. The resulting skill set is $\mathcal{S}\!=\!\mathcal{S}^{\texttt{SI}}\!\cup\!\mathcal{S}^{\texttt{SC}}$ providing a structured task prior for the next stage.

\begin{figure}
	\begin{center}
           \includegraphics[width=1.0\linewidth]{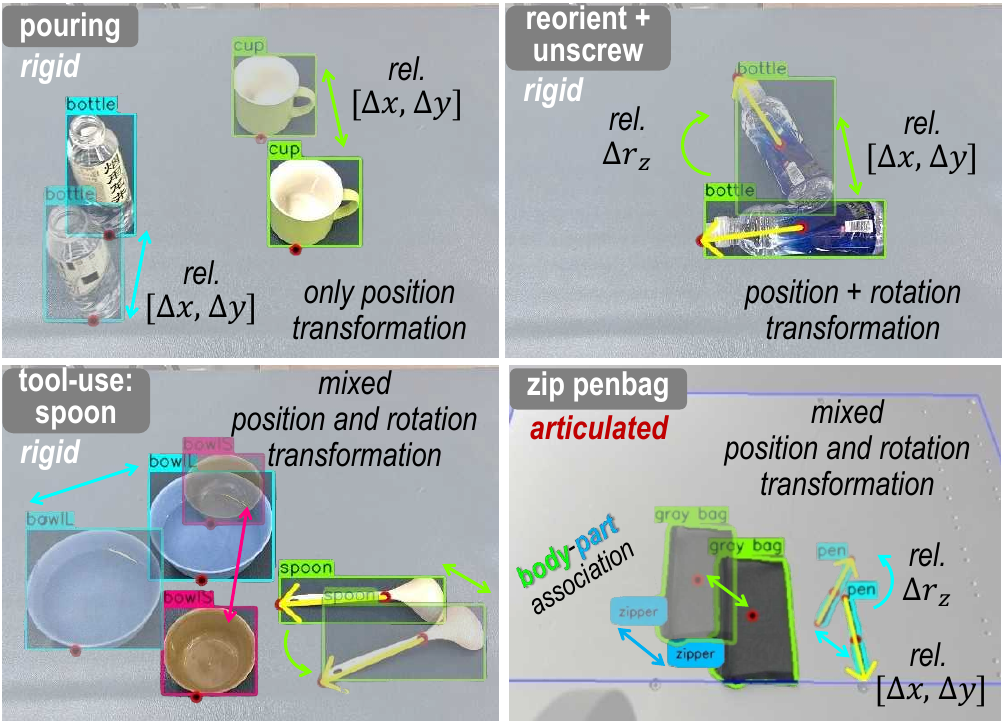}
	\vspace{-20pt}
	\caption{Illustrations of representative points for target objects in four tasks including \texttt{pouring}, \texttt{reorient+unscrew}, \texttt{tool-use:spoon}, and \texttt{zip penbag}. The first three tasks involve only rigid objects, while the last task involves an articulated object. These anchor points will be used to calculate the change in object position and orientation (not always required). }
           \label{locationPose} 
	\vspace{-15pt}
	\end{center}
\end{figure}

\subsection{Vision-Language Anchored Adaptation}\label{VLA}
After task-aware decomposition, VLBiMan++ adapts state-conditioned skills to novel object and scene states through vision-language grounding and object-centric geometric reasoning. Rather than estimating a complete physical scene state, we identify task-relevant objects and extract compact anchors that preserve the functional relationships encoded by the one-shot demonstration. The process consists of \textit{VLM-Based Scene Understanding}, \textit{Object-Centric Geometric Anchoring}, and \textit{Object-State-Aware Adaptation}.

\subsubsection{VLM-Based Scene Understanding}
Given the task description $\mathcal{T}$, we extract task-relevant prompts ${p_k}_{k=1}^{K}$ and query pretrained vision-language and vision foundation models, such as Florence-2 \cite{xiao2024florence} and SAM2 \cite{ravi2025sam}, to obtain semantic object masks $\mathbf{M}^{\texttt{2D}}_k$ from the current observation $\mathcal{O}^{\texttt{new}}$ of each target object $o_k$. The masks establish explicit correspondence between language-described targets and their visual instances, allowing the system to disambiguate relevant objects in cluttered scenes. We further reconstruct a 3D dense point cloud $\mathcal{P}^{\texttt{3D}}$ using calibrated stereo matching algorithms \cite{xu2023iterative, xu2025igev} and lift each mask to an object-level point cloud $\mathcal{P}^{\texttt{3D}}_k$. This provides the semantic and geometric information required for subsequent adaptation without task-specific detectors, CAD models, or additional training. The strong out-of-distribution perception capability of modern foundation models also makes this grounding procedure naturally robust to changes in unexpected illumination and distractors.

\begin{algorithm}[t]\small 
	\caption{Orientation Estimation for Rigid Objects}
	\begin{algorithmic}[1]
	\Require Binary object mask $\mathbf{M}^\texttt{2D} \in \{0,1\}^{H \times W}$
	\Ensure Orientation angle $\theta \in [0,360)$ (degrees)
	\State Extract object contour $C = \{(x_i,y_i)\}_{i=1}^N$ from $\mathbf{M}^\texttt{2D}$
	\State Compute the centroid point $(\bar{x}, \bar{y})$ using image moments: $\bar{x} = \frac{1}{N}\sum\nolimits_{i=1}^N x_i$, $\bar{y} = \frac{1}{N}\sum\nolimits_{i=1}^N y_i$
	\State Calculate second-order central moments: $\mu_{20} = \frac{1}{N}\sum (x_i-\bar{x})^2$, $\mu_{02} = \frac{1}{N}\sum (y_i-\bar{y})^2$, $\mu_{11} = \frac{1}{N}\sum (x_i-\bar{x})(y_i-\bar{y})$
	\State Construct covariance matrix: $\Sigma = \begin{bmatrix} \mu_{20} & \mu_{11} \\ \mu_{11} & \mu_{02} \end{bmatrix}$
	\State Compute eigenvalues $\lambda_1 > \lambda_2$ and eigenvectors $\mathbf{v}_1, \mathbf{v}_2$ of $\Sigma$
	\State Obtain principal axis direction $\mathbf{a} = (a_x, a_y) = \mathbf{v}_1$
	\State Project contour points onto principal axis: 
		\begin{equation*}
		p_i = (x_i-\bar{x})a_x + (y_i-\bar{y})a_y, \quad \forall i \in [1,N]
		\end{equation*}
	\State Identify endpoints: $e_\text{max} = \arg\max_i p_i, \; e_\text{min} = \arg\min_i p_i$
	\State Calculate perpendicular width $w_j$ within radius $r$ around each $e_j$
	\State Determine front endpoint: $e_\text{front} \gets (w_\text{max} < w_\text{min}) ? e_\text{max} : e_\text{min}$
	\State Adjust axis direction: $\mathbf{a} \gets (\mathbf{a} \cdot (e_\text{front} - (\bar{x},\bar{y})) < 0) ? -\mathbf{a} : \mathbf{a}$
	\State Final orientation angle: $\theta = \left(\arctan2(a_y, a_x) \times \frac{180}{\pi}\right) \bmod 360$
	\State \Return $\theta$
	\end{algorithmic}
	\label{algA}
\end{algorithm}

\subsubsection{Object-Centric Geometric Anchoring}
Instead of estimating a complete 6-DoF pose for every object \cite{lin2024sam, wen2024foundationpose} or generating grasp proposals \cite{fang2020graspnet, fang2023anygrasp}, VLBiMan++ constructs compact task-relevant anchors according to object type. Examples of defined anchor points are shown in Fig.~\ref{locationPose}.

For \textit{\textbf{rigid objects}}, we use either the 2D mask centroid or a task-specific point on the foremost object--table contact boundary as the representative point. Given the corresponding 3D positions $\mathbf{p}^{\texttt{demo}}_k$ and $\mathbf{p}^{\texttt{new}}_k$, the positional variation is $\Delta\mathbf{x}_k=\mathbf{p}^{\texttt{new}}_k-\mathbf{p}^{\texttt{demo}}_k$. For orientation-sensitive objects, we further extract the principal axis from the 2D mask using the image-moment method in Alg.~\ref{algA}, and compute $\Delta\theta_k=\angle(\mathbf{v}^{\texttt{new}}_k, \mathbf{v}^{\texttt{demo}}_k)$. Specifically, the first-order moments determine the mask centroid, while the second-order image moments \cite{chaumette2004image, kotoulas2007accurate} define a covariance matrix whose dominant eigenvector provides the principal axis. The resulting $(\Delta\mathbf{x}_k,\Delta\theta_k)$ is used to transform the demonstrated grasp or alignment pose into the current robot frame through the calibrated hand--eye transformation. Importantly, orientation estimation is invoked only when the manipulation is direction-sensitive. And symmetric objects with task-prescribed configurations require only positional anchoring.

For \textit{\textbf{articulated objects}}, the complete object may undergo configuration changes while the manipulation is still governed by a specific functional part. We therefore establish an explicit \textit{body-part association} \cite{zhou2023body, zhou2024bpjdet} between the semantic object and its task-relevant component. Let $o_k$ denote the complete articulated object and $r_k$ denote its relevant part. Once $r_k$ is identified, the latter is treated as the effective manipulation target, while the corresponding local geometry is regarded as a rigid entity for geometric anchoring. The same rigid-object procedure above, including position and orientation estimation when required, can therefore be applied to the local functional part without recovering the full articulated configuration. An example of an articulated penbag is shown in Fig.~\ref{locationPose}.

\begin{algorithm}[t]\small 
	\caption{Anchors Extraction for Deformable Objects}  
	\begin{algorithmic}[1]
	\Require Binary object mask $\mathbf{M}^\texttt{2D} \in \{0,1\}^{H \times W}$, Object prior $c \in \{\texttt{Rectangular}, \texttt{Linear}\}$, Anchors number $K$
	\Ensure Ordered anchors sequence $\mathcal{P} = \{\mathbf{p}_1, \dots, \mathbf{p}_K\}$
	
	\Statex \textbf{Phase 1: Morphological Preprocessing}
	\State $\mathbf{M}^\texttt{2D}_{\text{closed}} \gets \text{MorphClose}(\mathbf{M}^\texttt{2D}, \text{kernel})$
	\State $\mathbf{M}^\texttt{2D}_{\text{smooth}} \gets \text{Threshold}(\text{GaussianBlur}(\mathbf{M}^\texttt{2D}_{\text{closed}}, \sigma))$
	
	\Statex \textbf{Phase 2 \& 3: Category-Specific Extraction}
	\If{$c == \texttt{Rectangular}$}
	    \State Extract largest external contour $\mathcal{C}$ from $\mathbf{M}^\texttt{2D}_{\text{smooth}}$
	    \State Compute convex hull $\mathcal{H} \gets \text{ConvexHull}(\mathcal{C})$
	    \State $\mathcal{V} \gets \text{ApproxPolyDP}(\mathcal{H}, \epsilon)$ \textcolor{gray}{\Comment{Approximate to 4 vertices}}
	    \State Compute centroid $\bar{\mathbf{v}} = \frac{1}{4}\sum_{\mathbf{v} \in \mathcal{V}} \mathbf{v}$
	    \State Sort $\mathbf{v}_i \in \mathcal{V}$ by polar angle $\arctan2(\mathbf{v}_{i,y} - \bar{\mathbf{v}}_y, \mathbf{v}_{i,x} - \bar{\mathbf{v}}_x)$ 
	    \State Align starting node: shift $\mathcal{V}$ s.t. $\mathbf{v}_1 = \arg\min_{\mathbf{v}} \|\mathbf{v}\|_2$
	    \State $\mathcal{P} \gets \mathcal{V}$ \textcolor{gray}{\Comment{Output 4 ordered corners}}
	
	\ElsIf{$c == \texttt{Linear}$}
	    \State $\mathbf{S} \gets \text{Skeletonize}(\mathbf{M}^\texttt{2D}_{\text{smooth}})$
	    \State Build undirected graph $G=(V, E)$ from skeleton pixels $\mathbf{S}$
	    \State $\mathcal{J} \gets \{v \in V \mid \text{deg}(v) \ge 3\}$ \textcolor{gray}{\Comment{Identify junction nodes}}
	    
	    \For{each junction cluster $J \subseteq \mathcal{J}$}
	        \State Compute cluster centroid $\mathbf{c}_J$, identify exit branches $E_J$
	        \State Obtain lookahead vector $\vec{d}_e$ for each $e \in E_J$
	        \State Greedy pair $(e_i, e_j)$ if internal angle $\langle \vec{d}_{e_i}, \vec{d}_{e_j} \rangle \to \pi$
	        \State Update $G \gets (G \setminus J) \cup \{(e_i, e_j)\}$ \textcolor{gray}{\Comment{Decouple crossings}}
	    \EndFor
	    
	    \State Find principal path $\Pi = \{\pi_1, \dots, \pi_M\}$ on $G$  
	    \State Densify \& smooth $\tilde{\Pi} \gets \text{GaussianFilter}(\text{Interpolate}(\Pi), \sigma_{\text{path}})$  
	    
	    \State Compute cumulative arc length $L(m) = \sum_{j=1}^{m} \|\tilde{\pi}_j - \tilde{\pi}_{j-1}\|_2$ 
	    \For{$k \gets 1$ to $K$}
	        \State Target length $l_k = \frac{k-1}{K} \cdot L(|\tilde{\Pi}|)$
	        \State $\mathbf{p}_k \gets \tilde{\pi}_{\bar{m}}$, $\bar{m} = \arg\min_m |L(m) - l_k|$  \textcolor{gray}{\Comment{Quantile}} 
	    \EndFor
	\EndIf
	
	\State \Return $\mathcal{P}$
	\end{algorithmic}
	\label{algB}
\end{algorithm}  

\begin{figure}
	\begin{center}
           \includegraphics[width=1.0\linewidth]{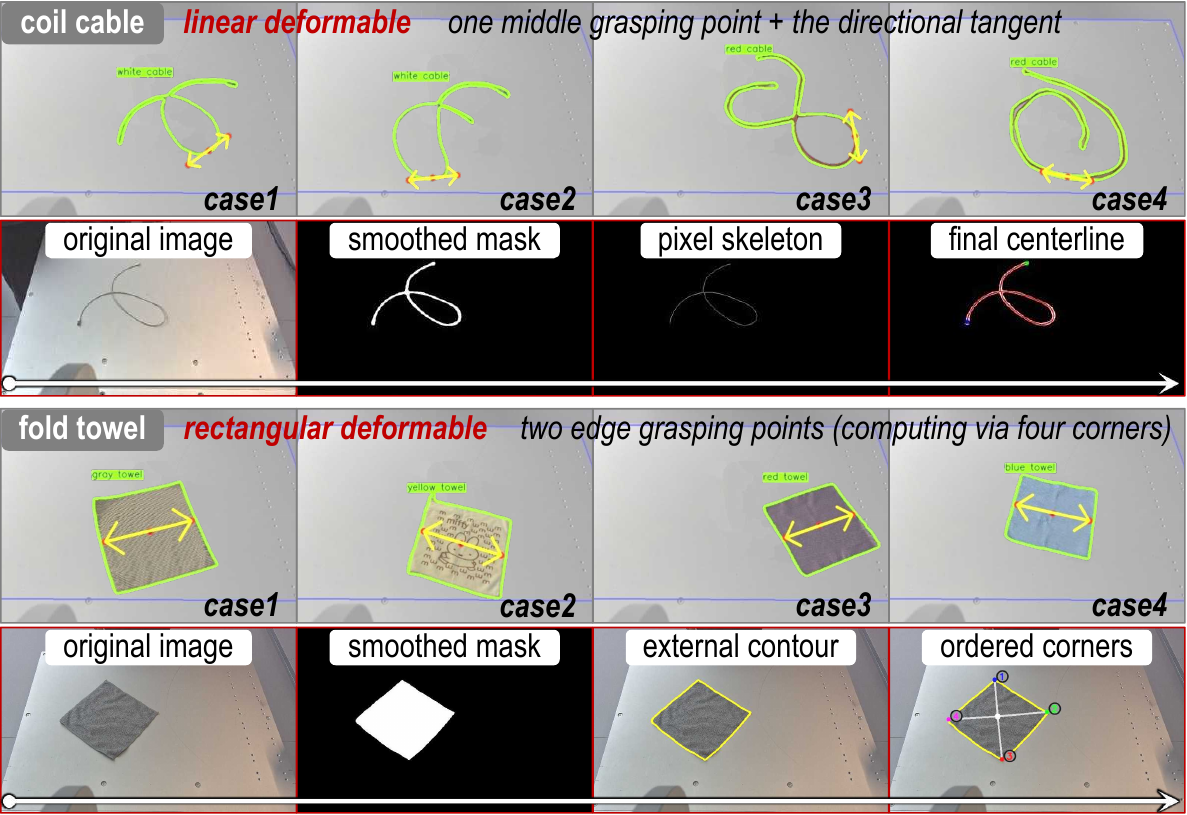}
	\vspace{-20pt}
	\caption{Illustrations of extracted various anchor points for deformable objects in two tasks including \texttt{coil cable} (top) and \texttt{fold towel} (bottom). These can serve as a supplement to Alg.~\ref{algB}. We present the computational results for four test cases of each task. In addition, we also present intermediate results for handling linear or rectangular deformable objects.}
           \label{nonRigidObject} 
	\vspace{-15pt}
	\end{center}
\end{figure}

For \textit{\textbf{deformable objects}}, complete pose estimation is insufficient to represent the task-relevant geometry. Given the inherent complexity of modeling high-dimensional deformations, we focus on two prevalent topological categories: approximately \textit{rectangular} objects (\textit{e.g.}, towels) and \textit{linear} objects (\textit{e.g.}, cables). Instead of estimating a complete deformation state, we characterize the visible object by leveraging category-specific geometric priors to extract stable manipulation anchors. After morphologically smoothing the binary mask $\mathbf{M}^\texttt{2D}$ of a segmented deformable object to eliminate sensor noise \cite{bradski2000}, the algorithm routes the extraction process based on its topology. For \textit{rectangular} objects, we enforce shape convexity on the external contour and employ a polygonal approximation to isolate four primary vertices. A centroid-based polar sorting mechanism is then applied to guarantee a consistent, ambiguity-free cyclic ordering of corners. For \textit{linear} objects, we extract the medial-axis skeleton and construct an undirected topological graph. To robustly handle self-intersections, we cluster high-degree junction nodes and geometrically decouple them by greedily pairing exit branches based on their macroscopic tangent vectors. The continuous principal curve is subsequently recovered via Eulerian path tracing or Dijkstra's diameter search \cite{dijkstra2022}, enabling precise fractional sampling (\textit{i.e.}, quantile locations) along the smoothed trajectory. This topology-aware extraction procedure retains critical task-relevant shape information with high efficiency, which is summarized in Alg.~\ref{algB}. And a broader visualization of anchors extraction results are presented in Fig.~\ref{nonRigidObject}.

\subsubsection{Object-State-Aware Adaptation}
Let $s_k$ denote the task-relevant object state $o_k$. Given the demonstrated state $s_k^{\texttt{demo}}$ and the current state $s_k^{\texttt{new}}$, we define the state variation as $\Delta s_k\!=\!s_k^{\texttt{new}}\!-\!s_k^{\texttt{demo}}$, where the subtraction denotes the representation-specific difference, such as translational displacement, angular deviation, articulation change, or boundary-anchor variation. VLBiMan++ selectively updates only state-conditioned skills:
\begin{equation}
	\widetilde{\mathcal{M}}^{\texttt{SI}}_i = \mathcal{M}^{\texttt{SI}}_i, \quad \widetilde{\mathcal{M}}^{\texttt{SC}}_j = \mathcal{A}(\mathcal{M}^{\texttt{SC}}_j,\Delta s_k),
	\label{state_adaptation}
\end{equation}
where $\mathcal{A}(\cdot)$ denotes the corresponding anchor-based adaptation operation. For rigid objects, $\Delta s_k$ is primarily reflected by $(\Delta\mathbf{x}_k,\Delta\theta_k)$ and size variation. For articulated objects, it additionally contains the state of the relevant movable part. For deformable objects, it is represented by the variation of the extracted boundary anchors. Thus, VLBiMan++ preserves invariant manipulation structure while adapting only the components affected by object-state changes, and passes the resulting skills to the trajectory composition stage.

\subsection{Autonomous Trajectory Composition}\label{ATC}
After state-aware adaptation, VLBiMan++ composes the state-invariant and state-conditioned skills according to the temporal structure of the demonstration. Since direct concatenation may introduce reachability or collision issues under novel object states, we apply progressive IK refinement and dynamic collision compensation, followed by closed-loop re-observation and recomposition during execution.

\subsubsection{Progressive IK Refinement}
For the initial grasping stage, directly solving IK from the current end-effector pose to the adapted grasp pose may yield unreachable or unsafe motions. VLBiMan++ therefore approaches the target through interpolated Cartesian waypoints:
\begin{equation}
	\begin{aligned}
	&\mathbf{q}^{(n+1)}=\texttt{IK}(\mathbf{T}_{g}^{(n)}), \\
	&\mathbf{T}_{g}^{(n)}=\texttt{SplineInterp}(\mathbf{T}_{\texttt{start}}, \mathbf{T}_{\texttt{goal}}, n ),
	\label{IKrefine}
	\end{aligned}
\end{equation}
where $\mathbf{T}_{\texttt{start}}$ is the currently observed end-effector pose, $\mathbf{T}_{\texttt{goal}}$ is the adapted grasp pose obtained from Sec.~\ref{VLA}, and $n$ is the interpolation density. This converts a large reaching motion into a sequence of smaller and more feasible IK subproblems  \cite{chitta2012moveit, schulman2014motion}, improving reachability and providing intermediate states for object-state re-evaluation. If the object is displaced during this process, $\mathbf{T}_{\texttt{goal}}$ is recomputed from the current observation. Otherwise, it remains unchanged.

\subsubsection{Dynamic Collision Compensation}
To reduce premature contact during pre-grasping, we modify the adapted target position with proximal and vertical compensation:
\begin{equation}
	\widetilde{\mathbf{x}}^{\texttt{goal}} = \mathbf{x}^{\texttt{goal}} + \delta_{\texttt{base}}\mathbf{u}_{\parallel} + \delta_z\mathbf{u}_{z},
	\label{DynaColl}
\end{equation}
where $\delta_{\texttt{base}}$ and $\delta_z$ denote the proximal and vertical compensation terms, respectively, and $\mathbf{u}_{\parallel}$ and $\mathbf{u}_{z}$ specify the corresponding Cartesian directions. The proximal compensation increases clearance along the horizontal approach direction, while the vertical compensation provides additional separation for top-down reaching. After trajectory synthesis, a one-time physical replay is used to identify unintended object or inter-arm collisions, after which the corresponding waypoints are adjusted before reuse. Thus, collision-related corrections are incorporated into the reusable skill rather than repeatedly optimized during subsequent deployments.

\subsubsection{Closed-Loop Re-observation and Re-composition}
Real-world execution may deviate from the planned state because of external perturbations or robot--object interactions. VLBiMan++ therefore monitors the task-relevant object state and triggers re-adaptation when $| s_k^{\texttt{exec}} - s_k^{\texttt{plan}} | > \epsilon$, where $s_k^{\texttt{exec}}$ and $s_k^{\texttt{plan}}$ denote the observed and planned states, respectively. The $\epsilon$ denotes the stability tolerance used to distinguish effective object displacement from perceptual fluctuations. Once triggered, the target is re-grounded and the affected state-conditioned skills are re-instantiated:
\begin{equation}
	\widetilde{\mathcal{M}}^{\texttt{SC}}_j \leftarrow \mathcal{A} \left( \mathcal{M}^{\texttt{SC}}_j, s_k^{\texttt{exec}} - s_k^{\texttt{demo}} \right), \quad 
	\widetilde{\mathcal{M}}^{\texttt{SI}}_i \leftarrow \mathcal{M}^{\texttt{SI}}_i .
	\label{recomposition}
\end{equation}
The resulting trajectory is then recomposed while preserving all unaffected invariant skills. This selective closed-loop update enables VLBiMan++ to withstand repeated object relocation and execution-induced displacement without retraining, following the cycle \textit{Observe} $\rightarrow$ \textit{Ground} $\rightarrow$ \textit{Adapt} $\rightarrow$ \textit{Compose} $\rightarrow$ \textit{Execute} $\rightarrow$ \textit{Re-observe}. This design preserves the reusable task prior extracted from a single demonstration while enabling sustained operation under changing object states and external disturbances, forming the basis for deployment-level generalization investigated in this work.



\section{Experiments}
We aim to answer four key questions: (1) How effectively does VLBiMan++ perform diverse bimanual manipulation tasks without retraining (Sec.~\ref{overall})? (2) How far can a single demonstration generalize across tasks, objects, scenes, and robot embodiments (Sec.~\ref{generalization})? (3) How robustly can VLBiMan++ sustain manipulation under dynamic disturbances and long-term closed-loop execution (Sec.~\ref{robustness})? (4) How do individual components contribute to the performance, generalization, and robustness of the system (Sec.~\ref{analysis})? We evaluate VLBiMan++ on diverse real-world bimanual manipulation tasks using heterogeneous robotic platforms.

\begin{table*}[t]  
	\centering
	\setlength{\tabcolsep}{0.7pt}
	\caption{Detailed statistics on all bimanual manipulation tasks, including 10 previous tasks and 9 newly added tasks. In the \textbf{Dual-Arm Platform(s)}, P1 and P2 represent the task has been instantiated on a contralateral robot and a humanoid robot, respectively. In the \textbf{Object Properties}, Rig., Art., and Def. represent rigid, articulated, and deformable objects, respectively. Among them, deformable objects require Alg.~\ref{algB} to extract anchor points. And the \textbf{Orientation Estimation} is based on Alg.~\ref{algA}.}
	\vspace{-8pt}
	\begin{tabular}{c||c|c|c|c|c|c | c|c|c|c||c|c|c|c|c|c | c|c|c}
	\Xhline{1.2pt}
	~ & \multicolumn{6}{c|}{\textit{\textbf{\cellcolor{green!10}primary bimanual tasks/skills}}} & \multicolumn{4}{c||}{\textit{\textbf{\cellcolor{green!30}long-horizon multi-stage tasks}}} & \multicolumn{6}{c|}{\textit{\textbf{\cellcolor{red!10}new rearrangement tasks}}} & \multicolumn{3}{c}{\textit{\textbf{\cellcolor{red!30}new non-rigid targets}}} \\
	\cline{2-20}
	\textbf{\makecell{Task\\Types\\\&\\Names}}
		& \rotatebox[origin=c]{80}{\texttt{plugpen}} & \rotatebox[origin=c]{80}{\texttt{inserting}} & \rotatebox[origin=c]{80}{\texttt{unscrew}} & \rotatebox[origin=c]{80}{\texttt{pouring}} & \rotatebox[origin=c]{80}{\texttt{pressing}} & \rotatebox[origin=c]{80}{\texttt{handover}}
 		& \rotatebox[origin=c]{80}{\texttt{\makecell{reorient\\+unscrew}}} & \rotatebox[origin=c]{80}{\texttt{\makecell{unscrew\\+pouring}}} & \rotatebox[origin=c]{80}{\texttt{\makecell{tool-use\\scoop}}} & \rotatebox[origin=c]{80}{\texttt{\makecell{tool-use\\funnel}}}
		& \rotatebox[origin=c]{80}{\texttt{flatting}} & \rotatebox[origin=c]{80}{\texttt{reorient}} & \rotatebox[origin=c]{80}{\texttt{flipping}} & \rotatebox[origin=c]{80}{\texttt{upright}}
		& \rotatebox[origin=c]{80}{\texttt{\makecell{place \\bottle\_mug}}} & \rotatebox[origin=c]{80}{\texttt{\makecell{place \\fork\_spoon}}}
 		& \rotatebox[origin=c]{80}{\texttt{\makecell{zip \\penbag}}} & \rotatebox[origin=c]{80}{\texttt{\makecell{coil \\cable}}} & \rotatebox[origin=c]{80}{\texttt{\makecell{fold \\towel}}} \\
	\Xhline{0.8pt} 
	\textbf{\makecell{Dual-Arm\\Platform(s)}} & P1$|$P2 & P1$|$P2 & P1$|$P2 & P1$|$P2 & P1 & P1$|$P2 & P1 & P1$|$P2 & P1 & P1 & P2 & P2 & P2 & P2 & P2 & P2 & P2 & P2 & P2 \\ \hline
	\textbf{\makecell{Object\\Number}} & 2 & 2 & 1 & 2 & 2 & 1 & 1 & 2 & 3 & 3 & 1 & 1 & 1 & 1 & 2 & 2 & 2 & 1 & 1 \\ \hline
	\textbf{\makecell{Object\\Names\\/States}} & \makecell{pen\\body\\+cap} & \makecell{pen\\+cup} & \makecell{upward\\bottle} & \makecell{upward\\bottle\\+mug} & \makecell{cup+\\pump\\bottle} & \makecell{spoon \\or \\shovel} 
		& \makecell{lying\\bottle} & \makecell{upward\\bottle\\+mug}  & \makecell{spoon+\\bowl\_1+\\bowl\_2} & \makecell{funnel\\+bottle\\+mug}
		& \makecell{upside\\down\\bottle} & \makecell{lying\\bottle} & \makecell{upside\\down\\mug} & \makecell{lying\\mug} & \makecell{bottle\\+mug} & \makecell{fork+\\spoon} 
		& \makecell{penbag\\+pen} & \makecell{linear\\cable} & \makecell{rectan\\-gular\\towel} \\ \hline
	\textbf{\makecell{Object\\Properties}} & Art. & Rig. & Rig. & Rig. & Rig. & Rig. & Rig. & Rig. & Rig. & Rig. & Rig. & Rig. & Rig. & Rig. & Rig. & Rig. & Art. & Def. & Def. \\ \hline
	\textbf{\makecell{Orientation\\Estimation?}} & \ding{51}\ding{51} & \ding{51}\ding{55} & \ding{55} & \ding{55}\ding{55} & \ding{55}\ding{55} & \ding{51} & \ding{51} & \ding{55}\ding{55} & \ding{51}\ding{55}\ding{55} & \ding{51}\ding{55}\ding{55} 
 		& \ding{55} & \ding{51} & \ding{55} & \ding{51} & \ding{51}\ding{51} & \ding{51}\ding{51} & \ding{51}\ding{51} & \ding{55} & \ding{55} \\
	\Xhline{1.2pt}
	\end{tabular}
	\label{tabA}
	\vspace{-5pt}
\end{table*}

\begin{figure*}[t]
    \centering
    \begin{minipage}{.515\textwidth}
		\centering
		\includegraphics[width=\columnwidth]{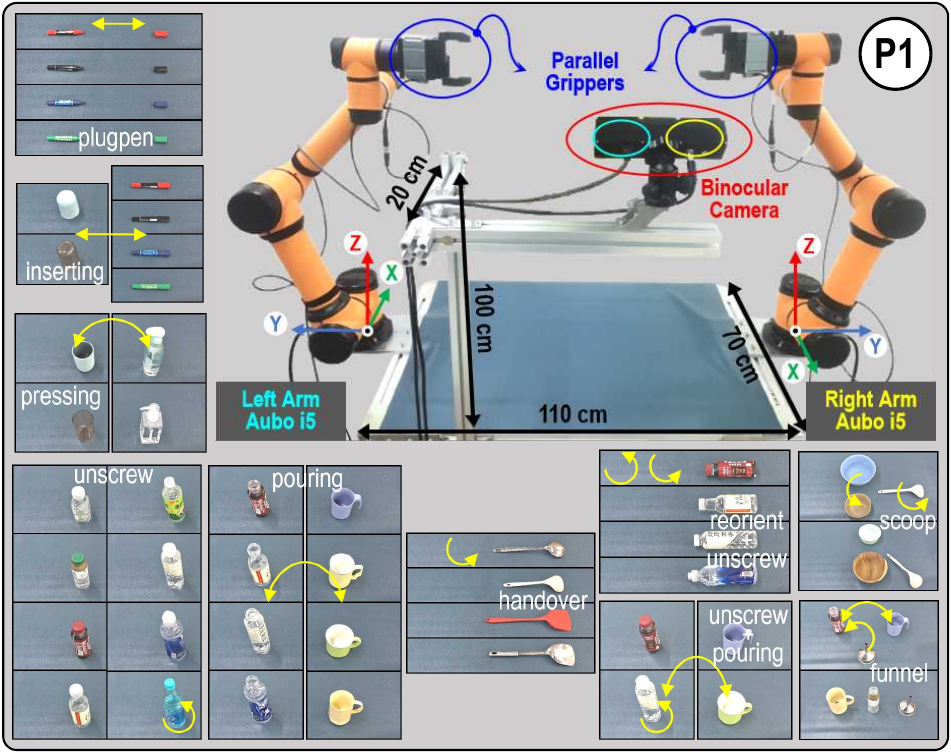}
		\vspace{-20pt}
		\caption{The contralateral dual-arm platform (\textbf{P1}), and all manipulated object assets involved in ten previously defined bimanual tasks.}
		\label{hardwareA}
    \end{minipage}
    \hspace{0.005cm}
    \begin{minipage}{.468\textwidth}
		\centering
		\includegraphics[width=\columnwidth]{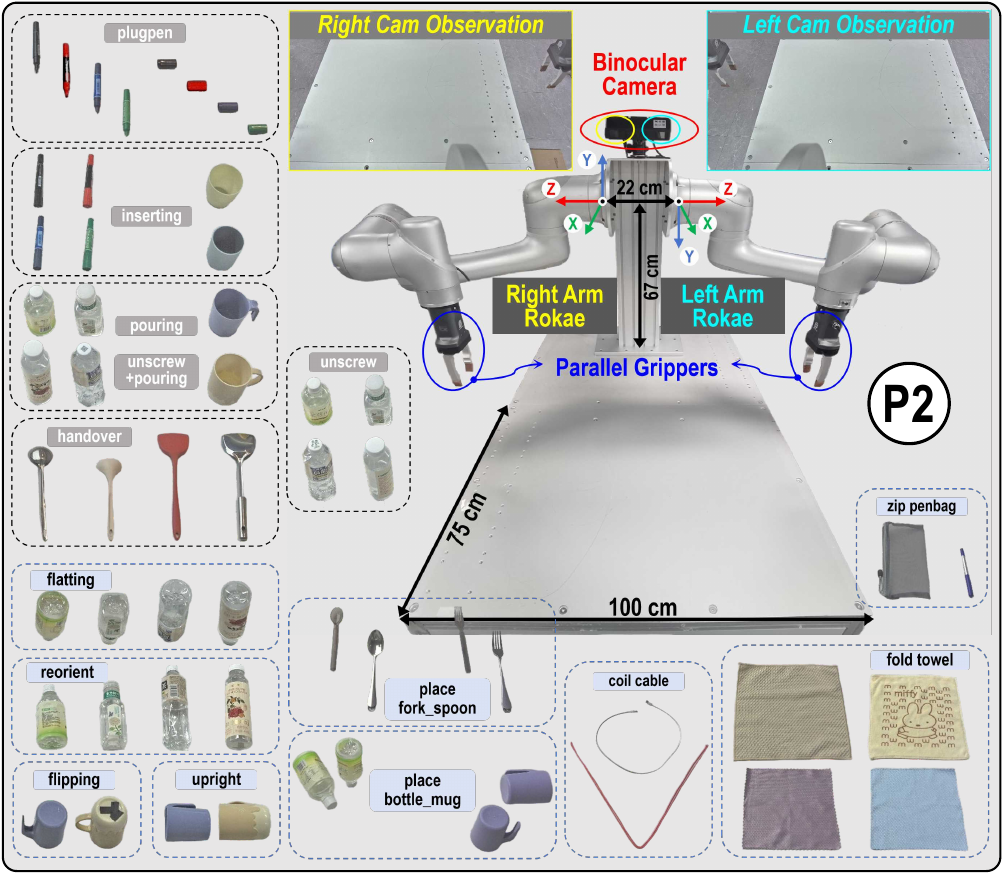}
		\vspace{-20pt}
		\caption{The humanoid platform (\textbf{P2}), and all manipulated object assets involved in six previous and nine newly defined bimanual tasks.}
		\label{hardwareB}
    \end{minipage}
    \vspace{-10pt}
\end{figure*}

\subsection{Experimental Setup}

\subsubsection{Tasks and Robotic Platforms}
\begin{figure*}[t]
	\begin{center}
           \includegraphics[width=0.9\linewidth]{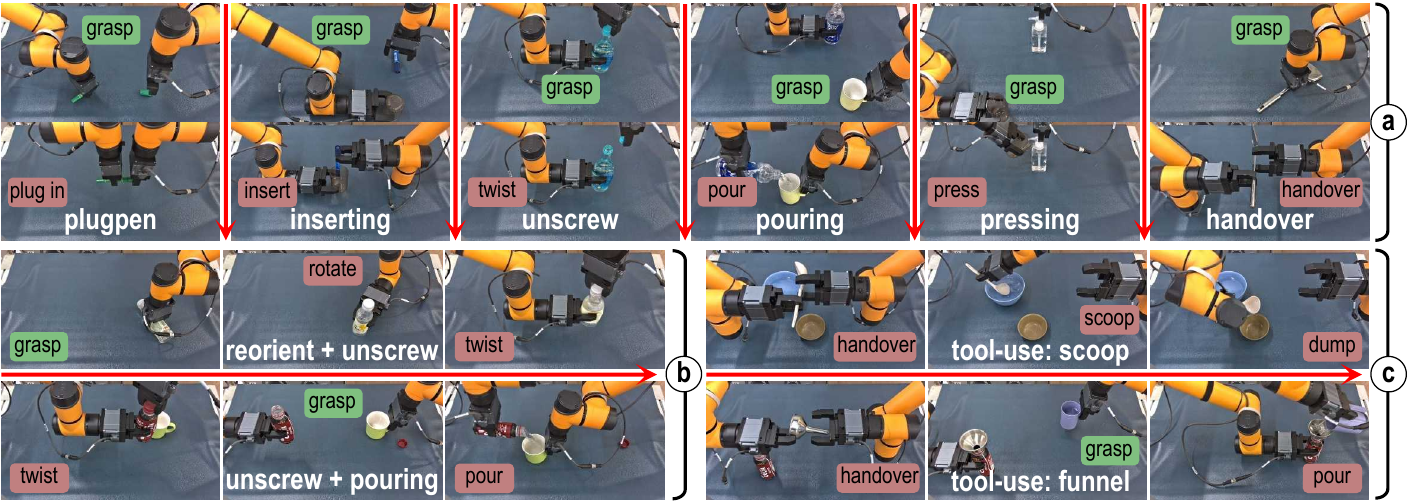}
	\vspace{-10pt}
	\caption{Visualizations of ten tasks executed on the contralateral dual-arm robots platform \textbf{P1}. They are designed to validate different aspects, including \textbf{(a)} six dual-arm primary skills, \textbf{(b)} combination of basic skills for two long-horizon tasks, and \textbf{(c)} exploration of two multi-stage tool-use tasks.}
           \label{visualization}
	\vspace{-15pt}
	\end{center}
\end{figure*}

\begin{figure*}[t]
	\begin{center}
           \includegraphics[width=\linewidth]{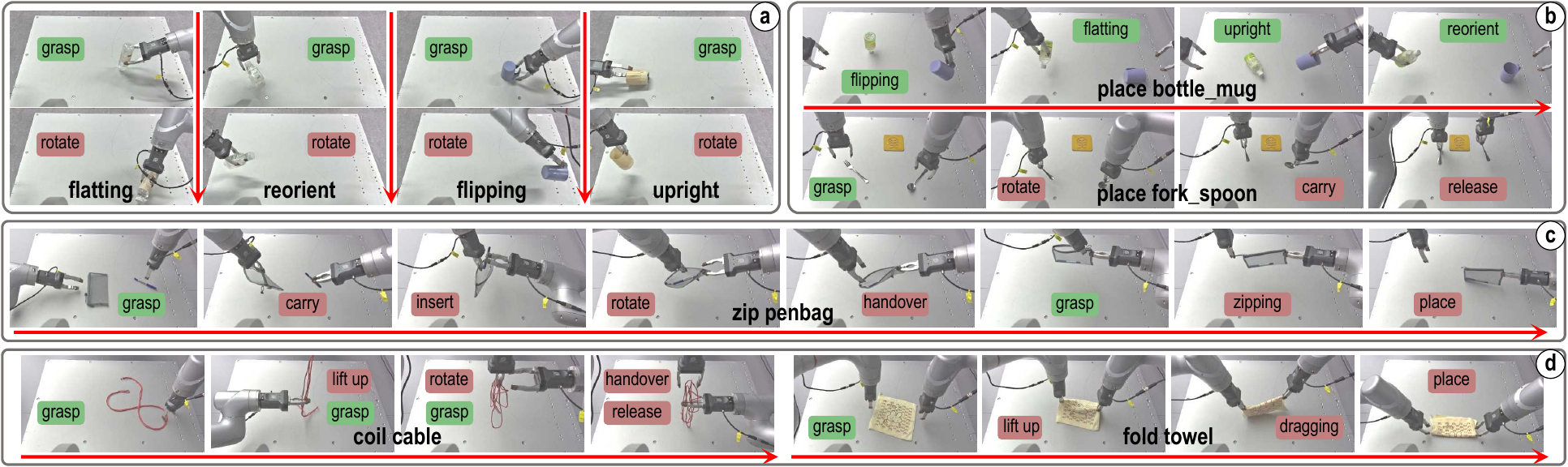}
	\vspace{-20pt}
	\caption{Visualizations of newly added nine tasks executed on the humanoid-style dual-arm platform \textbf{P2}. They are defined to further confirm the universality of VLBiMan++, including \textbf{(a)} four left-arm or right-arm rearrangement skills, \textbf{(b)} combination of new basic skills for two long-horizon rearrangement tasks, \textbf{(c)} a complex task involving articulated bodies, \textbf{(d)} two dual-arm tasks involving linear or rectangular deformable objects.}
           \label{visualizationPlus}
	\vspace{-15pt}
	\end{center}
\end{figure*}

We evaluate VLBiMan++ on a substantially expanded suite of real-world bimanual manipulation tasks covering diverse skill compositions, object states, and environmental conditions (refer Tab.~\ref{tabA}). The core benchmark retains the ten tasks from our conference version (refer Fig.~\ref{hardwareA} and Fig.~\ref{visualization}), including six primary bimanual tasks/skills\footnote{We have renamed the inherently dual-arm task \texttt{reorient} from the original conference version to \texttt{handover} to avoid confusion with the newly added single-arm (left or right) task \texttt{reorient}.} and four more challenging long-horizon or tool-use tasks. In this journal extension, we further introduce nine newly-defined tasks and object configurations to probe generalization beyond original rigid-object settings, including unseen object categories, articulated and deformable objects, cluttered scenes, and prolonged disturbance-aware execution (refer Fig.~\ref{hardwareB} and Fig.~\ref{visualizationPlus}). The experiments are conducted on two heterogeneous dual-arm platforms: a contralateral fixed-base setup equipped with two 6-DoF Aubo-i5 arms\footnote{https://www.aubo-cobot.com/public/i5product3} (880mm reach, denoted as \textbf{P1}), parallel grippers, and a third-person stereo camera, and a humanoid-style platform equipped with two 6-DoF Rokae xMate CR7 arms\footnote{https://www.rokae.com/en/product/show/545/xMateCR.html} (988mm reach, denoted as \textbf{P2}), parallel grippers, and a centrally mounted stereo camera. This setup enables evaluation across diverse tasks, object states, scenes, and robot embodiments. 


\subsubsection{Baselines and Evaluation Metrics}
We compare VLBiMan++ with representative training-free and data-efficient manipulation approaches, including Robot-ABC \cite{ju2024robo}, which combines affordance prediction with AnyGrasp \cite{fang2023anygrasp} for grasp generation, and ReKep \cite{huang2024rekep}, which leverages vision foundation models (SAM \cite{kirillov2023segment} and DINOv2 \cite{oquab2024dinov2}) and LLM-guided keypoint constraints for task execution (\textit{e.g.}, GPT-4o \cite{achiam2023gpt}). We further include one-shot manipulation methods such as Mechanisms \cite{mao2023learning} and MAGIC \cite{liu2025one}, adapted to our bimanual setting where applicable, as well as the enhanced ReKep+ variant with oracle initial grasp annotations for a stronger comparison. For all real-robot evaluations, we report task success rate over repeated trials under matched experimental conditions, considering both seen and unseen object instances as well as external disturbances when applicable. Additional analyses evaluate execution efficiency, robustness to clutter and repeated perturbations, and cross-embodiment transfer, providing a comprehensive assessment of VLBiMan++ across the proposed generalization boundary.

\begin{table*}[t]  
	\centering
	\setlength{\tabcolsep}{0.5pt}
	\caption{Quantitative comparison results of success rates on six primary bimanual tasks/skills and four long-horizon multi-stage tasks under the \textbf{in-distribution (ID)} setting (new placements + \textbf{same objects}) or the \textbf{out-of-distribution (OOD)} setting (new placements + \textbf{novel instances}) using the contralateral dual-arm robot (\textbf{P1}).}
	\vspace{-8pt}
	\begin{tabular}{c|c|c|cccccc|c|cccc|c}
	\Xhline{1.2pt}
	~ & ~ & ~ & \multicolumn{7}{c|}{\textit{\textbf{\cellcolor{green!10}six primary bimanual tasks/skills}}} & \multicolumn{5}{c}{\textit{\textbf{\cellcolor{green!30}four long-horizon multi-stage tasks}}} \\
	\cline{4-15}
	\makecell{Test\\Setting} & \makecell{Dyna\\-mic\\Inter-\\ference} & \makecell{Manipulation\\Method}
		& \rotatebox[origin=c]{45}{\texttt{plugpen}} & \rotatebox[origin=c]{45}{\texttt{inserting}} & \rotatebox[origin=c]{45}{\texttt{unscrew}} & \rotatebox[origin=c]{45}{\texttt{pouring}} & \rotatebox[origin=c]{45}{\texttt{pressing}} & \rotatebox[origin=c]{45}{\texttt{handover}}
 		& \makecell{~\\Average\\Success\\Rate} 
 		& \rotatebox[origin=c]{45}{\texttt{\makecell{reorient\\+unscrew}}} & \rotatebox[origin=c]{45}{\texttt{\makecell{unscrew\\+pouring}}} & \rotatebox[origin=c]{45}{\texttt{\makecell{tool-use\\scoop}}} & \rotatebox[origin=c]{45}{\texttt{\makecell{tool-use\\funnel}}}
 		& \makecell{~\\Average\\Success\\Rate} \\
	\Xhline{0.8pt} 
	\multirow{12}{*}{ID} & \multirow{6}{*}{No} & Mechanisms 
		& 11/25 & 09/25 & 05/25 & 05/25 & 07/25 & 03/25 & \cellcolor{gray!15}26.7\% 
		& 05/25 & 04/25 & 02/25 & 01/25 & \cellcolor{gray!15}12.0\%  \\
 	~ & ~ & MAGIC 
		& 16/25 & 15/25 & 10/25 & 10/25 & 09/25 & 07/25 & \cellcolor{gray!15}44.7\% 
		& 09/25 & 08/25 & 04/25 & 03/25 & \cellcolor{gray!15}24.0\%  \\
	~ & ~ & Robot-ABC 
		& 14/25 & 10/25 & 09/25 & 07/25 & 08/25 & 06/25 & \cellcolor{gray!15}36.0\% 
		& 06/25 & 06/25 & 03/25 & 03/25 & \cellcolor{gray!15}18.0\%  \\
	~ & ~ & ReKep 
		& 14/25 & 11/25 & 10/25 & 12/25 & 10/25 & 08/25 & \cellcolor{gray!15}43.3\% 
		& 07/25 & 08/25 & 05/25 & 03/25 & \cellcolor{gray!15}23.0\%  \\
	~ & ~ & ReKep+
		& 19/25 & 18/25 & 13/25 & 17/25 & 17/25 & 11/25 & \cellcolor{gray!15}63.3\% 
		& 11/25 & 10/25 & 07/25 & 06/25 & \cellcolor{gray!15}34.0\%  \\
	~ & ~ & \textbf{VLBiMan++}
		& 25/25 & 23/25 & 20/25 & 21/25 & 20/25 & 19/25 & \textbf{\cellcolor{gray!15}85.3\%}  
		& 15/25 & 15/25 & 12/25 & 10/25 & \textbf{\cellcolor{gray!15}52.0\%}  \\
	\cline{2-15}
	~ & \multirow{6}{*}{Yes} & Mechanisms 
		& 05/25 & 05/25 & 03/25 & 02/25 & 04/25 & 01/25 & \cellcolor{gray!15}13.3\% 
		& 01/25 & 02/25 & 00/25 & 00/25 & \cellcolor{gray!15}3.0\%  \\
 	~ & ~ & MAGIC 
		& 09/25 & 09/25 & 05/25 & 04/25 & 06/25 & 04/25 & \cellcolor{gray!15}24.7\% 
		& 04/25 & 03/25 & 04/25 & 01/25 & \cellcolor{gray!15}12.0\%  \\
	~ & ~ & Robot-ABC 
		& 07/25 & 06/25 & 04/25 & 03/25 & 05/25 & 02/25 & \cellcolor{gray!15}18.0\% 
		& 02/25 & 02/25 & 02/25 & 02/25 & \cellcolor{gray!15}9.0\%  \\
	~ & ~ & ReKep 
		& 10/25 & 06/25 & 06/25 & 04/25 & 05/25 & 03/25 & \cellcolor{gray!15}22.7\% 
		& 06/25 & 05/25 & 03/25 & 02/25 & \cellcolor{gray!15}16.0\%  \\
	~ & ~ & ReKep+
		& 12/25 & 10/25 & 09/25 & 08/25 & 09/25 & 09/25 & \cellcolor{gray!15}38.0\% 
		& 08/25 & 08/25 & 05/25 & 03/25 & \cellcolor{gray!15}24.0\%  \\
	~ & ~ & \textbf{VLBiMan++}
		& 19/25 & 16/25 & 19/25 & 18/25 & 17/25 & 15/25 & \textbf{\cellcolor{gray!15}69.3\%}  
		& 12/25 & 11/25 & 09/25 & 06/25 & \textbf{\cellcolor{gray!15}38.0\%}  \\
	\Xhline{0.6pt} 
	\multirow{12}{*}{OOD} & \multirow{6}{*}{No} & Mechanisms
		& 06/25 & 05/25 & 02/25 & 01/25 & 04/25 & 01/25 & \cellcolor{gray!15}12.7\% 
		& 01/25 & 02/25 & 00/25 & 00/25 & \cellcolor{gray!15}3.0\% \\  
	~ & ~ & MAGIC
		& 11/25 & 10/25 & 05/25 & 05/25 & 06/25 & 04/25 & \cellcolor{gray!15}27.3\% 
		& 05/25 & 04/25 & 01/25 & 01/25 & \cellcolor{gray!15}11.0\% \\  
	~ & ~ & Robot-ABC 
		& 11/25 & 09/25 & 03/25 & 02/25 & 07/25 & 04/25 & \cellcolor{gray!15}24.0\% 
		& 04/25 & 02/25 & 00/25 & 01/25 & \cellcolor{gray!15}7.0\% \\  
	~ & ~ & ReKep 
		& 12/25 & 08/25 & 05/25 & 06/25 & 07/25 & 06/25 & \cellcolor{gray!15}29.3\% 
		& 05/25 & 04/25 & 01/25 & 00/25 & \cellcolor{gray!15}10.0\% \\  
	~ & ~ & ReKep+
		& 15/25 & 12/25 & 09/25 & 10/25 & 11/25 & 07/25 & \cellcolor{gray!15}42.7\% 
		& 07/25 & 06/25 & 04/25 & 02/25 & \cellcolor{gray!15}19.0\% \\  
	~ & ~ & \textbf{VLBiMan++}
		& 24/25 & 21/25 & 18/25 & 17/25 & 20/25 & 17/25 & \textbf{\cellcolor{gray!15}78.0\%} 
		& 12/25 & 11/25 & 10/25 & 08/25 & \textbf{\cellcolor{gray!15}41.0\%} \\  
	\cline{2-15}
	~ & \multirow{6}{*}{Yes} & Mechanisms
		& 03/25 & 01/25 & 00/25 & 00/25 & 02/25 & 00/25 & \cellcolor{gray!15}4.0\% 
		& 00/25 & 00/25 & 00/25 & 00/25 & \cellcolor{gray!15}0.0\% \\  
	~ & ~ & MAGIC
		& 05/25 & 04/25 & 03/25 & 01/25 & 04/25 & 01/25 & \cellcolor{gray!15}12.0\% 
		& 02/25 & 02/25 & 01/25 & 00/25 & \cellcolor{gray!15}5.0\% \\  
	~ & ~ & Robot-ABC 
		& 05/25 & 03/25 & 00/25 & 00/25 & 03/25 & 00/25 & \cellcolor{gray!15}7.3\%  
		& 01/25 & 00/25 & 01/25 & 00/25 & \cellcolor{gray!15}2.0\% \\  
	~ & ~ & ReKep 
		& 09/25 & 04/25 & 03/25 & 01/25 & 04/25 & 02/25 & \cellcolor{gray!15}15.3\% 
		& 03/25 & 03/25 & 00/25 & 00/25 & \cellcolor{gray!15}6.0\% \\  
	~ & ~ & ReKep+
		& 10/25 & 08/25 & 05/25 & 04/25 & 06/25 & 05/25 & \cellcolor{gray!15}25.3\% 
		& 06/25 & 04/25 & 01/25 & 01/25 & \cellcolor{gray!15}12.0\% \\  
	~ & ~ & \textbf{VLBiMan++}
		& 18/25 & 14/25 & 15/25 & 14/25 & 15/25 & 13/25 & \textbf{\cellcolor{gray!15}59.3\%} 
		& 08/25 & 09/25 & 05/25 & 02/25 & \textbf{\cellcolor{gray!15}24.0\%} \\  
	\Xhline{1.2pt}
	\end{tabular}
	\label{tabB}
	\vspace{-10pt}
\end{table*}

\begin{table*}[t]  
	\centering
	\setlength{\tabcolsep}{1pt}
	\caption{Quantitative comparison results of success rates on six new rearrangement tasks and three new non-rigid targets under the \textbf{in-distribution (ID)} setting (new placements + \textbf{same objects}) or the \textbf{out-of-distribution (OOD)} setting (new placements + \textbf{novel instances}) using the humanoid dual-arm robot (\textbf{P2}). The '---' means there are no extra new object instances.}
	\vspace{-8pt}
	\begin{tabular}{c|c|c|cccccc|c|ccc|c}
	\Xhline{1.2pt}
	~ & ~ & ~ & \multicolumn{7}{c|}{\textit{\textbf{\cellcolor{red!10}six new rearrangement tasks}}} & \multicolumn{4}{c}{\textit{\textbf{\cellcolor{red!30}three new non-rigid targets}}} \\
	\cline{4-14}
	\makecell{Test\\Setting} & \makecell{Dyna\\-mic\\Inter-\\ference} & \makecell{Manipulation\\Method}
		& \rotatebox[origin=c]{45}{\texttt{flatting}} & \rotatebox[origin=c]{45}{\texttt{reorient}} & \rotatebox[origin=c]{45}{\texttt{flipping}} & \rotatebox[origin=c]{45}{\texttt{upright}}
		& \rotatebox[origin=c]{45}{\texttt{\makecell{place \\bottle\_mug}}} & \rotatebox[origin=c]{45}{\texttt{\makecell{place \\fork\_spoon}}}
 		& \makecell{~\\Average\\Success\\Rate} 
 		& \rotatebox[origin=c]{45}{\texttt{\makecell{zip \\penbag}}} & \rotatebox[origin=c]{45}{\texttt{\makecell{coil \\cable}}} & \rotatebox[origin=c]{45}{\texttt{\makecell{fold \\towel}}}
 		& \makecell{~\\Average\\Success\\Rate} \\
	\Xhline{0.8pt} 
	\multirow{10}{*}{ID} & \multirow{6}{*}{No} & Mechanisms 
		& 04/25 & 04/25 & 02/25 & 01/25 & 01/25 & 01/25 & \cellcolor{gray!15}8.7\% 
		& 00/25 & 00/25 & 02/25 & \cellcolor{gray!15}2.7\%  \\
 	~ & ~ & MAGIC 
		& 06/25 & 05/25 & 03/25 & 02/25 & 01/25 & 02/25 & \cellcolor{gray!15}12.7\% 
		& 00/25 & 00/25 & 03/25 & \cellcolor{gray!15}4.0\%  \\
	~ & ~ & Robot-ABC 
		& 07/25 & 06/25 & 06/25 & 06/25 & 05/25 & 05/25 & \cellcolor{gray!15}23.3\% 
		& 01/25 & 00/25 & 05/25 & \cellcolor{gray!15}8.0\%  \\
	~ & ~ & ReKep+
		& 13/25 & 12/25 & 10/25 & 11/25 & 08/25 & 11/25 & \cellcolor{gray!15}43.3\% 
		& 04/25 & 00/25 & 08/25 & \cellcolor{gray!15}16.0\%  \\
	~ & ~ & \textbf{VLBiMan++}
		& 24/25 & 22/25 & 18/25 & 20/25 & 15/25 & 23/25 & \textbf{\cellcolor{gray!15}81.3\%}  
		& 12/25 & 10/25 & 20/25 & \textbf{\cellcolor{gray!15}56.0\%}  \\
	\cline{2-14}
	~ & \multirow{6}{*}{Yes} & Mechanisms 
		& 02/25 & 02/25 & 01/25 & 00/25 & 00/25 & 01/25 & \cellcolor{gray!15}4.0\% 
		& 00/25 & 00/25 & 00/25 & \cellcolor{gray!15}0.0\%  \\
 	~ & ~ & MAGIC 
		& 02/25 & 03/25 & 01/25 & 01/25 & 00/25 & 02/25 & \cellcolor{gray!15}6.0\% 
		& 00/25 & 00/25 & 00/25 & \cellcolor{gray!15}0.0\%  \\
	~ & ~ & Robot-ABC 
		& 05/25 & 05/25 & 03/25 & 03/25 & 01/25 & 04/25 & \cellcolor{gray!15}14.0\% 
		& 01/25 & 00/25 & 02/25 & \cellcolor{gray!15}4.0\%  \\
	~ & ~ & ReKep+
		& 08/25 & 08/25 & 06/25 & 05/25 & 03/25 & 08/25 & \cellcolor{gray!15}25.3\% 
		& 02/25 & 00/25 & 05/25 & \cellcolor{gray!15}9.3\%  \\
	~ & ~ & \textbf{VLBiMan++}
		& 15/25 & 14/25 & 13/25 & 13/25 & 11/25 & 15/25 & \textbf{\cellcolor{gray!15}54.0\%}  
		& 06/25 & 02/25 & 13/25 & \textbf{\cellcolor{gray!15}28.0\%}  \\
	\Xhline{0.6pt} 
	\multirow{10}{*}{OOD} & \multirow{6}{*}{No} & Mechanisms
		& 04/25 & 04/25 & 01/25 & 01/25 & --- & 01/25 & \cellcolor{gray!15}8.8\% 
		& --- & 00/25 & 01/25 & \cellcolor{gray!15}2.0\% \\  
	~ & ~ & MAGIC
		& 07/25 & 06/25 & 03/25 & 04/25 & --- & 02/25 & \cellcolor{gray!15}17.6\% 
		& --- & 00/25 & 03/25 & \cellcolor{gray!15}6.0\% \\  
	~ & ~ & Robot-ABC 
		& 09/25 & 08/25 & 04/25 & 06/25 & --- & 03/25 & \cellcolor{gray!15}24.0\% 
		& --- & 00/25 & 04/25 & \cellcolor{gray!15}8.0\% \\  
	~ & ~ & ReKep+
		& 12/25 & 10/25 & 06/25 & 09/25 & --- & 05/25 & \cellcolor{gray!15}33.6\% 
		& --- & 00/25 & 06/25 & \cellcolor{gray!15}12.0\% \\  
	~ & ~ & \textbf{VLBiMan++}
		& 23/25 & 20/25 & 15/25 & 18/25 & --- & 20/25 & \textbf{\cellcolor{gray!15}60.8\%} 
		& --- & 06/25 & 14/25 & \textbf{\cellcolor{gray!15}40.0\%} \\  
	\cline{2-14}
	~ & \multirow{6}{*}{Yes} & Mechanisms
		& 02/25 & 01/25 & 00/25 & 00/25 & --- & 00/25 & \cellcolor{gray!15}2.4\% 
		& --- & 00/25 & 00/25 & \cellcolor{gray!15}0.0\% \\  
	~ & ~ & MAGIC
		& 04/25 & 03/25 & 02/25 & 02/25 & --- & 01/25 & \cellcolor{gray!15}9.6\% 
		& --- & 00/25 & 00/25 & \cellcolor{gray!15}0.0\% \\  
	~ & ~ & Robot-ABC 
		& 05/25 & 04/25 & 02/25 & 02/25 & --- & 01/25 & \cellcolor{gray!15}11.2\% 
		& --- & 00/25 & 00/25 & \cellcolor{gray!15}0.0\% \\  
	~ & ~ & ReKep+
		& 07/25 & 06/25 & 03/25 & 04/25 & --- & 02/25 & \cellcolor{gray!15}19.2\% 
		& --- & 00/25 & 01/25 & \cellcolor{gray!15}2.0\% \\  
	~ & ~ & \textbf{VLBiMan++}
		& 16/25 & 12/25 & 09/25 & 11/25 & --- & 15/25 & \textbf{\cellcolor{gray!15}50.4\%} 
		& --- & 01/25 & 08/25 & \textbf{\cellcolor{gray!15}18.0\%} \\  
	\Xhline{1.2pt}
	\end{tabular}
	\label{tabC}
	\vspace{-10pt}
\end{table*}

\subsection{Overall Performance of VLBiMan++}\label{overall}
We first evaluate the overall capability of VLBiMan++ via a substantially expanded real-world tests. In addition to the ten tasks inherited from VLBiMan (mainly on the contralateral platform \textbf{P1}), the journal version further introduces six object-rearrangement tasks and three non-rigid manipulation tasks (mainly on the humanoid platform \textbf{P2}), covering a broader range of task structures and object states. For each task, we evaluate both in-distribution (ID) and out-of-distribution (OOD) settings, including novel object instances and spatial configurations, and report the corresponding success rates in Tab.~\ref{tabB} and Tab.~\ref{tabC}. Unless otherwise specified, each setting is evaluated over 25 real-robot trials.

\subsubsection{General Bimanual Manipulation}
As summarized in Tab.~\ref{tabB} and Tab.~\ref{tabC}, VLBiMan++ maintains the strong performance of the original VLBiMan while substantially broadening the scope of evaluated manipulation behaviors. The system consistently achieves reliable execution on foundational bimanual tasks as well as long-horizon and tool-use tasks, and further extends to the newly introduced rearrangement and non-rigid manipulation settings. This indicates that the task-aware decomposition and object-state-aware adaptation introduced in VLBiMan++ are not restricted to a small set of rigid-object behaviors, but can support more diverse task structures and object conditions. Notably, the same one-shot demonstration is reused as the task prior across the corresponding task variations, while only state-dependent components are adapted to the current scene. The resulting trajectories preserve the essential functional relationships of the demonstrated behaviors, including precise grasping, object alignment, inter-arm coordination, and multi-stage skill composition. The consistently high performance across both ID and OOD settings therefore demonstrates that VLBiMan++ improves not only task coverage, but also the practical reusability of a single human demonstration.

\subsubsection{Comparison with Existing Methods}
We further compare VLBiMan++ with representative state-of-the-art methods under the same real-world evaluation protocol. As shown in  Tabs.~\ref{tabB} and~\ref{tabC}, VLBiMan++ achieves the most competitive overall success rates while requiring only a single human demonstration and no policy retraining. In contrast, data-driven imitation and VLA policies \cite{zhao2023learning, chi2023diffusion, black2025pi0, pertsch2025fast} typically require substantial task-specific robot demonstrations and additional optimization when adapting to new tasks or objects, whereas existing training-free modular methods such as ReKep \cite{huang2024rekep} and Robot-ABC \cite{ju2024robo} must regenerate task-dependent grasps or trajectories for each new configuration. The adapted another two baselines Mechanisms \cite{mao2023learning} and MAGIC \cite{liu2025one} originally designed for single-arm tasks also cannot handle these bimanual tasks well, revealing the non-trivial nature of dual-arm coordination. This comparison highlights an important distinction in evaluation philosophy: VLBiMan++ is designed for one-shot, training-free generalization, rather than maximizing performance through additional task-specific data. Within this setting, its ability to reuse invariant skills, selectively adapt state-dependent components, and operate across novel objects and scenes provides a favorable balance between success rate, demonstration efficiency, and deployment cost. The advantage becomes more pronounced as task complexity, object diversity, and environmental variation increase, where repeatedly reconstructing complete manipulation trajectories becomes increasingly inefficient. Overall, these results demonstrate that VLBiMan++ preserves the effectiveness of VLBiMan while substantially extending its applicability to a broader range of bimanual manipulation problems, establishing a strong foundation for the subsequent analyses of task, object, scene, embodiment, and deployment generalization.

\subsection{Generalization across Different Configurations}\label{generalization}
We next evaluate how far the one-shot manipulation prior can be reused under increasingly diverse task, object, scene, and embodiment setttings. The results establish four intermediate levels of the generalization boundary introduced in VLBiMan++: task and skill composition, object-state variation, cluttered scenes, and heterogeneous robotic embodiments.

\subsubsection{Task and Skill Generalization}
We first investigate whether a single demonstration can provide reusable building blocks beyond the original task sequence. Starting from six primary bimanual tasks, VLBiMan++ reuses extracted atomic skills to construct long-horizon behaviors, novel combinations, and tool-use tasks. In particular, \texttt{reorient+unscrew}, \texttt{unscrew+pouring}, \texttt{place bottle\_mug} and \texttt{place fork\_spoon} compose previously acquired skills, while \texttt{tool-use spoon} and \texttt{tool-use funnel} further require their integration across multiple objects and stages. As shown in Tabs.~\ref{tabB} and~\ref{tabC}, VLBiMan++ maintains reliable performance across these increasingly complex settings without collecting additional demonstrations, demonstrating the transition from one demonstration to reusable skills and ultimately to new task compositions.

\begin{figure*}[t]
	\begin{center}
           \includegraphics[width=1.0\linewidth]{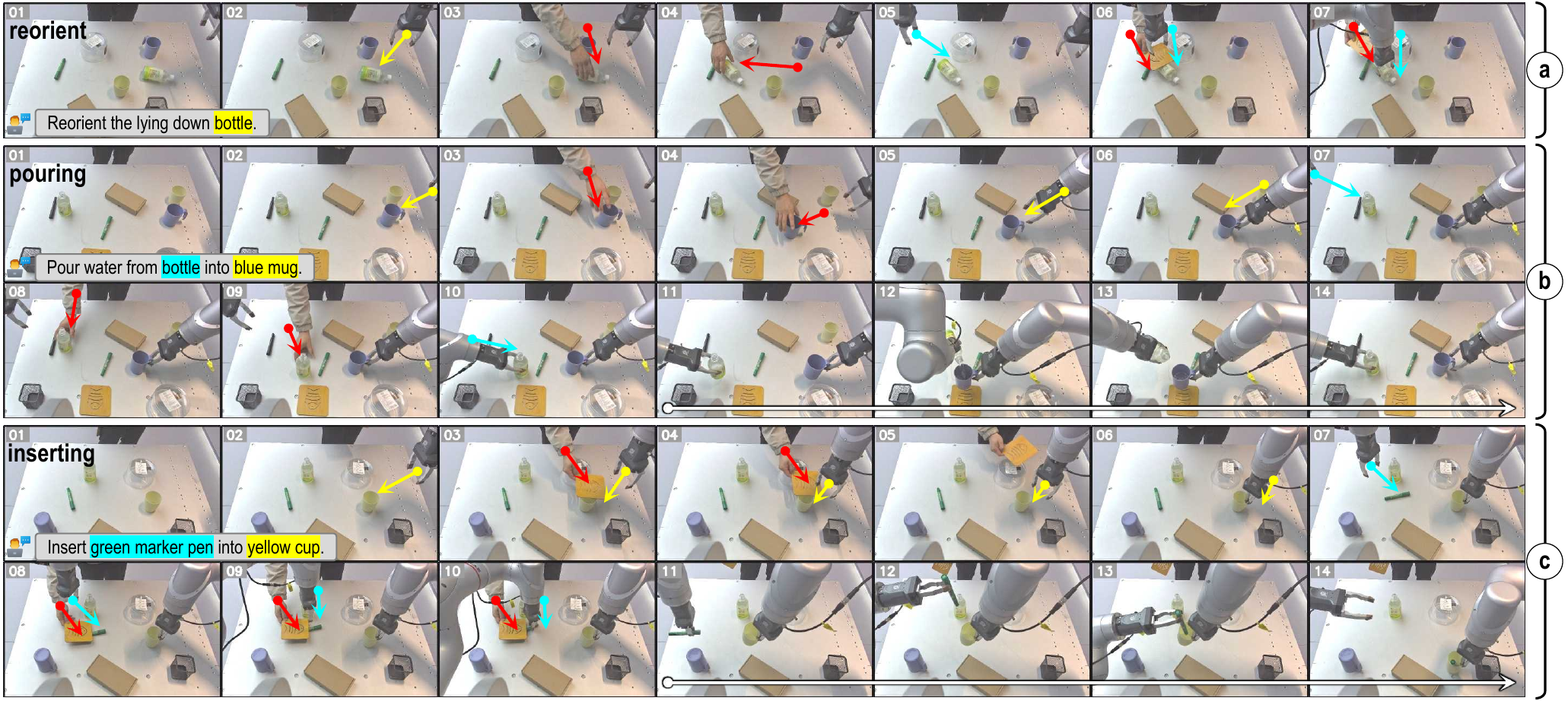}
	\vspace{-20pt}
	\caption{Visualizations of VLBiMan++'s robustness performance in cluttered scenarios on the platform \textbf{P2}. We selected three tasks for extensive evaluation: (\textbf{a}) \texttt{reorient}, (\textbf{b}) \texttt{pouring}, and (\textbf{c}) \texttt{inserting}. The \textcolor{red}{red} arrow indicates the direction of the manually moved or deliberately obscured object (interfering). The \textcolor{cyan}{cyan} arrow and \textcolor{olive}{yellow} arrow indicate the movement direction of the left and right robotic arms (chasing) respectively.}
           \label{clutterScene}
	\vspace{-15pt}
	\end{center}
\end{figure*}

\subsubsection{Unseen Object and Object-State Generalization}
We further examine the object-level generalization enabled by the \textit{Object-State-Aware Adaptation} module in Sec.~\ref{VLA}. For rigid objects, we evaluate unseen categories as well as variations in shape, size, position, and orientation. VLBiMan++ transfers the demonstrated manipulation structure by adapting only the task-relevant geometric anchors. We then extend the evaluation to articulated objects (\texttt{zip penbag}), where changes in movable-part configurations require adaptation of the corresponding body-part relationships. Finally, we consider non-rigid objects, including deformable structures (\texttt{coil cable} and \texttt{fold towel}) like cables and towels, for which VLBiMan++ employs topology-aware boundary anchors rather than a single rigid pose. Across these settings, the demonstrated task prior remains reusable while only the state-dependent components are modified, showing that the proposed framework extends beyond conventional rigid-instance transfer toward broader object-state generalization.

\subsubsection{Cluttered-Scene Generalization}
To assess scene-level generalization, we further evaluate VLBiMan++ in cluttered tabletop environments containing multiple distractors, semantic ambiguity, partial occlusion, and potential spatial interference. We conduct additional experiments on \texttt{reorient}, \texttt{pouring}, and \texttt{inserting}, with at least five irrelevant objects introduced into each scene (the platform is \textbf{P2}). Some distractors are visually or semantically similar to the target objects, while others may interfere with the manipulation path. The same vision-language grounding and adaptation pipeline is retained without task-specific modification. As shown in Fig.~\ref{clutterScene}, VLBiMan++ consistently identifies the language-specified targets, rejects irrelevant or similar objects, and maintains reliable manipulation despite partial occlusion and object relocation during execution. These results demonstrate that the object-centric anchors remain effective beyond clean tabletop configurations and that semantic grounding provides an important interface between cluttered visual observations and reusable manipulation skills.

\begin{table*}[h]  
	\centering
	\setlength{\tabcolsep}{2pt}
	\caption{Quantitative results of VLBiMan++'s success rates on \textbf{six transferred bimanual tasks} from the contralateral dual-arm robot (\textbf{P1}) to the humanoid dual-arm robot (\textbf{P2}) under the \textbf{ID} and \textbf{OOD} settings.}
	\vspace{-8pt}
	\begin{tabular}{c|c|cccccc|c|cccccc|c}
	\Xhline{1.2pt}
	\multirow{6}{*}{\makecell{Dynamic\\Interference}} & ~ & \multicolumn{7}{c|}{\textit{\textbf{\cellcolor{blue!10}ID setting: new placements + same objects}}} & \multicolumn{7}{c}{\textit{\textbf{\cellcolor{blue!30}OOD setting: new placements + novel instances}}} \\
	\cline{3-16}
	~ & \makecell{Dual-Arm\\Platform\\Type}
		& \rotatebox[origin=c]{60}{\texttt{plugpen}} & \rotatebox[origin=c]{60}{\texttt{inserting}} & \rotatebox[origin=c]{60}{\texttt{unscrew}} & \rotatebox[origin=c]{60}{\texttt{pouring}} & \rotatebox[origin=c]{60}{\texttt{handover}} & \rotatebox[origin=c]{60}{\texttt{\makecell{unscrew\\+pouring}}}
 		& \makecell{~\\Average\\Success\\Rate} 
		& \rotatebox[origin=c]{60}{\texttt{plugpen}} & \rotatebox[origin=c]{60}{\texttt{inserting}} & \rotatebox[origin=c]{60}{\texttt{unscrew}} & \rotatebox[origin=c]{60}{\texttt{pouring}} & \rotatebox[origin=c]{60}{\texttt{handover}} & \rotatebox[origin=c]{60}{\texttt{\makecell{unscrew\\+pouring}}}
 		& \makecell{~\\Average\\Success\\Rate} \\
	\Xhline{0.8pt} 
	\multirow{2}{*}{No} & Contralateral 
		& 19/20 & 18/20 & 16/20 & 17/20 & 15/20 & 12/20 & \cellcolor{gray!15}80.8\%  
		& 15/20 & 17/20 & 14/20 & 14/20 & 14/20 & 09/20 & \cellcolor{gray!15}69.2\% \\  
	~ & Humanoid
		& 20/20& 19/20 & 17/20 & 16/20 & 15/20 & 13/20 & \cellcolor{gray!15}83.3\%  
		& 16/20& 18/20 & 16/20 & 13/20 & 14/20 & 09/20 & \cellcolor{gray!15}71.7\% \\  
	\Xhline{0.6pt} 
	\multirow{2}{*}{Yes} & Contralateral
		& 16/20& 13/20 & 15/20 & 15/20 & 12/20 & 08/20 & \cellcolor{gray!15}65.8\%  
		& 14/20& 11/20 & 12/20 & 11/20 & 10/20 & 07/20 & \cellcolor{gray!15}54.2\% \\  
	~ & Humanoid
		& 16/20& 14/20 & 15/20 & 14/20 & 13/20 & 09/20 & \cellcolor{gray!15}67.5\%  
		& 12/20& 12/20 & 13/20 & 12/20 & 10/20 & 08/20 & \cellcolor{gray!15}55.8\% \\  
	\Xhline{1.2pt}
	\end{tabular}
	\label{tabD}
	\vspace{-10pt}
\end{table*}

\begin{figure}[t]
	\begin{center}
           \includegraphics[width=1.0\linewidth]{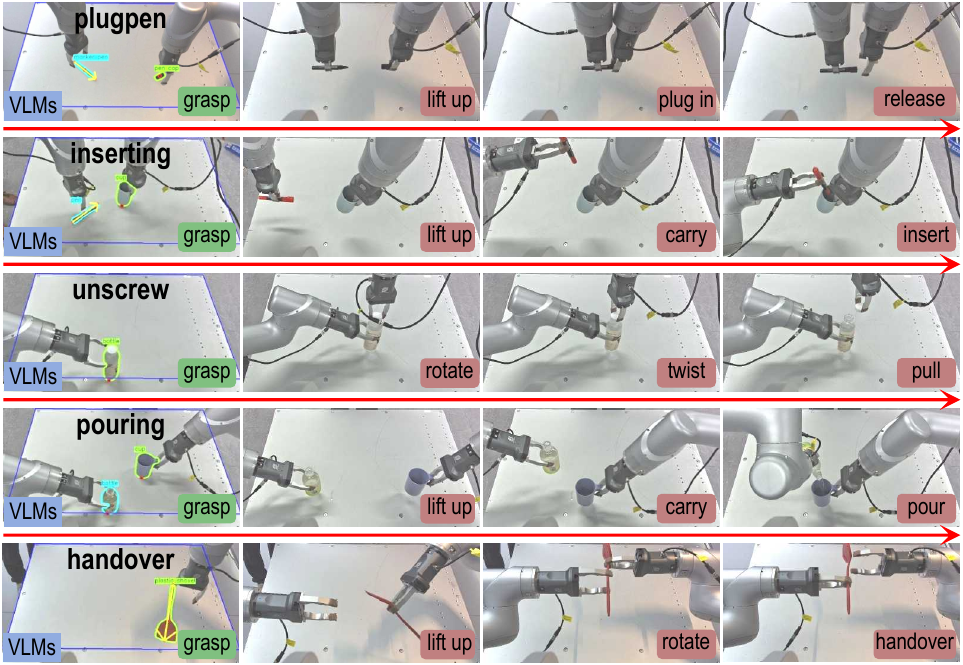}
	\vspace{-20pt}
	\caption{Visualizations of five cross-embodiment transferred primary bimaual tasks executed on the new humanoid-style robot (\textbf{P1} $\rightarrow$ \textbf{P2}).}
           \label{transferred}
	\vspace{-15pt}
	\end{center}
\end{figure}

\subsubsection{Cross-Embodiment Transfer}
Finally, we evaluate whether the task prior extracted from a single demonstration can be preserved when its physical realization changes across robotic embodiments. We transfer VLBiMan++ from the contralateral platform (\textbf{P1}) to a humanoid-style platform (\textbf{P2}) equipped with different robot arms, gripper specifications, and arm configurations. We evaluate \texttt{plugpen}, \texttt{inserting}, \texttt{unscrew}, \texttt{pouring}, \texttt{handover} and \texttt{unscrew+pouring}, including task configurations that are modified to accommodate the new embodiment. As summarized in Tab.~\ref{tabD}, VLBiMan++ maintains competitive performance on both seen and unseen objects, with additional robustness under external perturbations. These results indicate that the demonstration prior is not tied to the original robot's absolute configuration; instead, its task-relevant structure can be re-instantiated on a different embodiment. Qualitative executions in Fig.~\ref{transferred} further show that the humanoid platform can exploit more synchronized dual-arm motions, yielding smoother and more human-like execution.

\begin{table*}[t]  
	\centering
	\setlength{\tabcolsep}{1pt}
	\caption{Quantitative comparison results of success rates on \textbf{six primary bimanual skills/tasks} using the contralateral dual-arm robot (\textbf{P1}). All tasks faced the challenges of \textbf{dynamic interference} and \textbf{uneven illumination}.}
	\vspace{-8pt}
	\begin{tabular}{c|cccccc|c|cccccc|c}
	\Xhline{1.2pt}
	~ & \multicolumn{7}{c|}{\textit{\textbf{\cellcolor{blue!10}ID setting: new placements + same objects}}} & \multicolumn{7}{c}{\textit{\textbf{\cellcolor{blue!30}OOD setting: new placements + novel instances}}} \\
	\cline{2-15}
	\makecell{Manipulation\\Method}
		& \rotatebox[origin=c]{45}{\texttt{plugpen}} & \rotatebox[origin=c]{45}{\texttt{inserting}} & \rotatebox[origin=c]{45}{\texttt{unscrew}} & \rotatebox[origin=c]{45}{\texttt{pouring}} & \rotatebox[origin=c]{45}{\texttt{pressing}} & \rotatebox[origin=c]{45}{\texttt{handover}}
 		& \makecell{~\\Average\\Success\\Rate} 
		& \rotatebox[origin=c]{45}{\texttt{plugpen}} & \rotatebox[origin=c]{45}{\texttt{inserting}} & \rotatebox[origin=c]{45}{\texttt{unscrew}} & \rotatebox[origin=c]{45}{\texttt{pouring}} & \rotatebox[origin=c]{45}{\texttt{pressing}} & \rotatebox[origin=c]{45}{\texttt{handover}}
 		& \makecell{~\\Average\\Success\\Rate} \\
	\Xhline{0.8pt} 
	Mechanisms 
		& 01/20 & 01/20 & 00/20 & 01/20 & 01/20 & 00/20 & \cellcolor{gray!15}3.3\% 
		& 00/20 & 00/20 & 00/20 & 00/20 & 00/20 & 00/20 & \cellcolor{gray!15}0.0\% \\ 
 	MAGIC 
		& 03/20 & 04/20 & 02/20 & 02/20 & 02/20 & 01/20 & \cellcolor{gray!15}11.7\% 
		& 01/20 & 02/20 & 00/20 & 01/20 & 01/20 & 00/20 & \cellcolor{gray!15}4.2\% \\ 
	Robot-ABC 
		& 02/20 & 02/20 & 01/20 & 01/20 & 01/20 & 01/20 & \cellcolor{gray!15}6.7\% 
		& 00/20 & 00/20 & 00/20 & 00/20 & 00/20 & 00/20 & \cellcolor{gray!15}0.0\% \\ 
	ReKep 
		& 05/20 & 03/20 & 02/20 & 01/20 & 02/20 & 02/20 & \cellcolor{gray!15}12.5\% 
		& 03/20 & 01/20 & 01/20 & 01/20 & 01/20 & 00/20 & \cellcolor{gray!15}5.8\% \\ 
	ReKep+
		& 08/20 & 06/20 & 04/20 & 03/20 & 04/20 & 05/20 & \cellcolor{gray!15}25.0\% 
		& 05/20 & 02/20 & 03/20 & 02/20 & 03/20 & 01/20 & \cellcolor{gray!15}13.3\% \\ 
	\textbf{VLBiMan++}
		& 14/20 & 13/20 & 14/20 & 15/20 & 14/20 & 11/20 & \textbf{\cellcolor{gray!15}67.5\%}  
		& 13/20 & 11/20 & 11/20 & 10/20 & 12/20 & 10/20 & \textbf{\cellcolor{gray!15}55.8\%} \\  
	\Xhline{1.2pt}
	\end{tabular}
	\label{tabE}
	\vspace{-10pt}
\end{table*}

\subsection{Robustness and Long-Term Closed-Loop Deployment}\label{robustness}
Beyond static generalization, we further evaluate whether VLBiMan++ can sustain reliable manipulation under continuously changing execution conditions. We focus on three aspects: robustness to dynamic disturbances, stability during prolonged operation, and the effectiveness of the closed-loop re-observation and re-composition mechanism.

\subsubsection{Dynamic Disturbance Robustness}
We first evaluate robustness to external disturbances by varying the number and type of perturbations applied during execution. In addition to object relocation, we consider partial occlusion and uneven illumination, which jointly challenge visual grounding and geometric adaptation. Specifically, we conduct experiments on all six primary bimanual tasks using the platform \textbf{P1}, focusing on the ID evaluations without loss of generality. We define one interference as a scenario where each object involved in the task is disturbed once. Under this definition, we systematically vary the number of interferences from 0 to 5 and record the average task success rates, which are: 85.0\%, 70.0\%, 61.7\%, 56.7\%, 53.3\%, and 50.8\%, respectively. As the disturbance frequency increases, the success rate decreases monotonically, but the degradation becomes progressively smaller, indicating a diminishing marginal impact of repeated perturbations. This behavior suggests that the progressively updated object-relative anchors remain effective as the end-effector approaches the target. VLBiMan++ also exhibits strong robustness to adverse visual conditions. Under uneven illumination, the binary-mask-based geometric representation remains largely unaffected, while the underlying VLM/VFM perception maintains reliable object grounding. As summarized in Tab.~\ref{tabE}, the resulting performance degradation is substantially smaller than that of the compared various baselines, demonstrating that the proposed perception--adaptation pipeline is robust not only to geometric disturbances but also to appearance variations. Some qualitative examples of anti-interference on the platform \textbf{P2} can be found in Fig.~\ref{clutterScene}. 

\subsubsection{Super Long-Duration Execution}
\begin{figure*}[t]
	\begin{center}
           \includegraphics[width=1.0\linewidth]{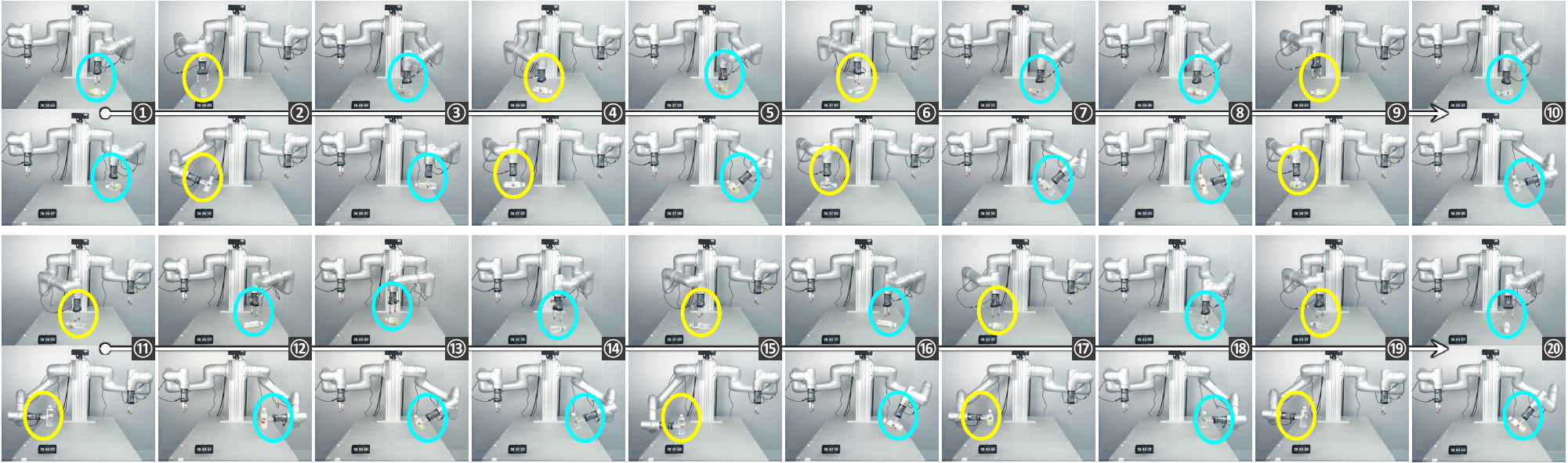}
	\vspace{-20pt}
	\caption{Video snapshots recorded from a third-person perspective for the super long-duration experiment of \texttt{reorient} on platform \textbf{P2}. It contains keyframes from \textbf{20} trials in which the left arm (\textcolor{cyan}{cyan} circle) or right arm (\textcolor{olive}{yellow} circle)  was used to upright a bottle lying down in an arbitrary orientation.}
           \label{longDuration}
	\vspace{-15pt}
	\end{center}
\end{figure*}

To further assess deployment stability, we conduct super long-duration tests using the platform \textbf{P2} in which the task \texttt{reorient} is executed continuously over \textbf{20} consecutive trials, with object states repeatedly randomized and external disturbances introduced throughout the process. Unlike conventional evaluations based on a limited number of independent trials, this setting examines whether small errors accumulate over time and whether the system can remain operational without retraining or manual re-initialization. The results show that VLBiMan++ maintains a stable success rate over prolonged execution, despite repeated object relocation and execution-induced disturbances. This stability is enabled by the selective reuse of state-invariant skills and repeated adaptation of only the affected components, preventing minor deviations from accumulating into a complete trajectory failure. The long-duration evaluation therefore provides evidence that the one-shot task prior remains reusable not only across isolated episodes, but also over sustained real-world deployment. We present the third-person video snapshots of this experiment in Fig.~\ref{longDuration}.

\subsubsection{Closed-Loop Re-observation and Re-composition}
\begin{figure}[t]
	\centering
	\includegraphics[width=\columnwidth]{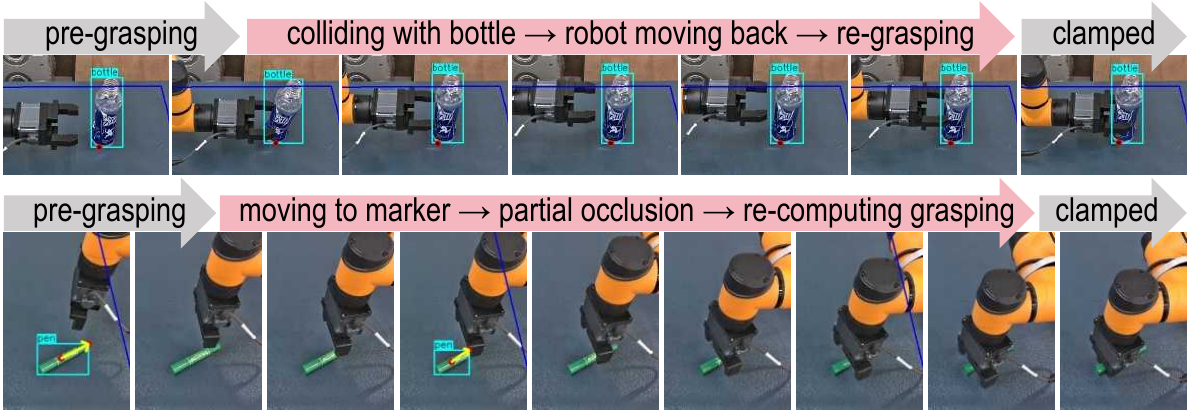}
	\vspace{-20pt}
	\caption{Examples of interesting findings about the closed-loop mechanism. \textit{\textbf{Top row}}: this case comes from the pre-grasping phase of \texttt{pouring}, where the left arm approaches and grasps the bottle. \textit{\textbf{Bottom row}}: this case comes from the pre-grasping phase of \texttt{inserting}, where the right arm approaches and grasps the marker from the top direction.}
	\label{findings}
	\vspace{-10pt}
\end{figure}

We finally isolate the contribution of the closed-loop mechanism introduced in Sec.~\ref{ATC} by comparing execution with and without re-observation and re-composition under repeated object perturbations. When the observed object state remains within the tolerance $\epsilon$ (\textit{e.g.}, 10 mm), VLBiMan++ continues the current trajectory without unnecessary replanning; once the discrepancy exceeds $\epsilon$, the system re-identifies the target, updates the corresponding state-conditioned skills, and recomposes the remaining trajectory. This mechanism leads to several practically important behaviors. During pre-grasping, the robot can dynamically refine its approach after an object is relocated. When the target is temporarily partially occluded, execution can continue using the temporally structured trajectory and resume adaptation once reliable visual grounding is recovered. Moreover, small object displacements during intermediate manipulation stages can often be tolerated without interrupting the task. These observations are consistent with the closed-loop execution traces shown in Fig.~\ref{findings} and demonstrate that VLBiMan++ is capable of selectively correcting its behavior rather than restarting the entire manipulation process.


\subsection{Ablation and In-Depth Analysis}\label{analysis}
We finally investigate why VLBiMan++ works and where its current limitations arise. We evaluate the robustness of object-centric representing points, the sensitivity to pre-grasp interpolation density, the contribution of individual modules, and the distribution of real-world failure cases.

\begin{figure*}[t]
	\begin{center}
           \includegraphics[width=1.0\linewidth]{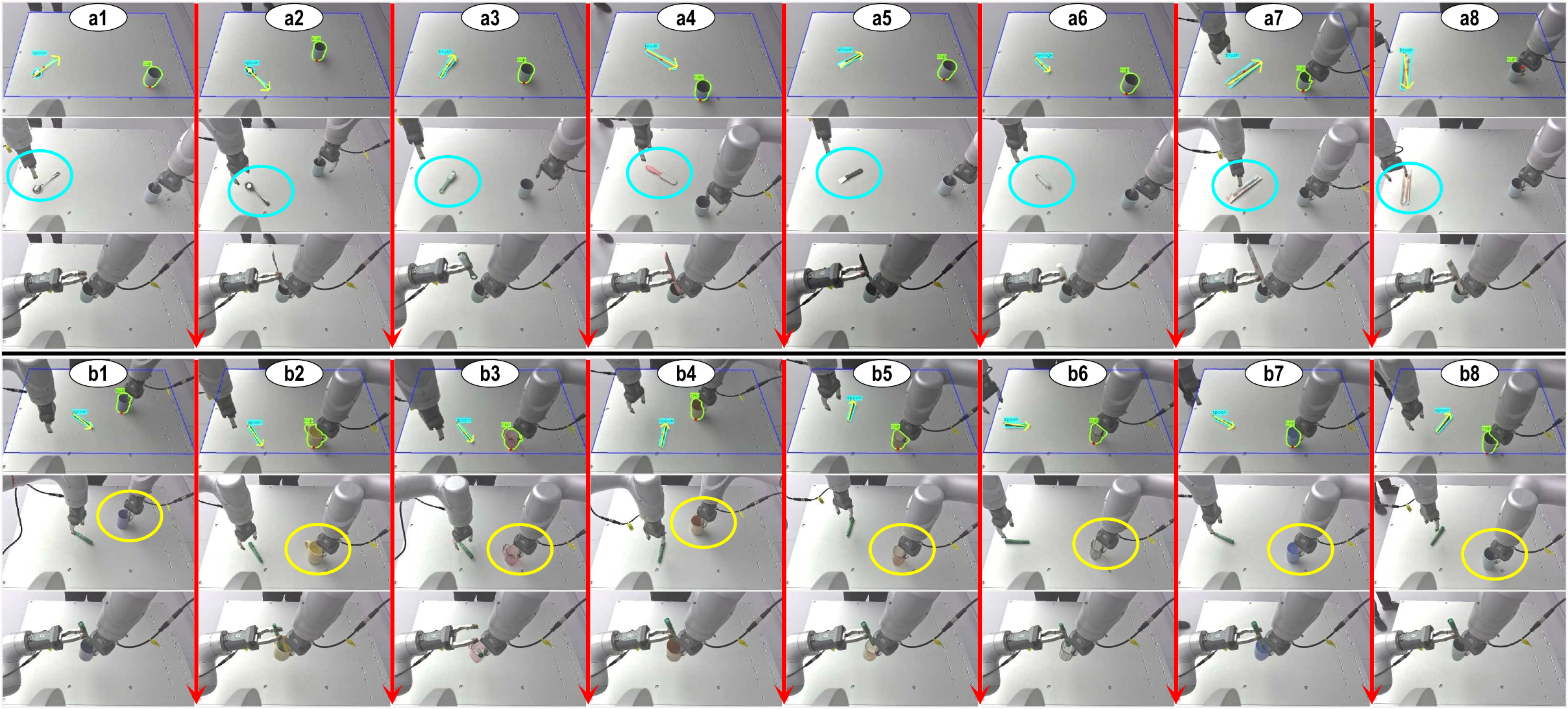}
	\vspace{-20pt}
	\caption{Taking the \texttt{inserting} task as an example, we can replace the marker pen held by the left arm with other rectangular objects that were completely different (\textit{e.g.}, \textit{spoon} in \textbf{a1/a2}, \textit{brush} in \textbf{a3/a4}, \textit{spatula} in \textbf{a5}, \textit{syringe} in \textbf{a6}, and \textit{toothbrush} in \textbf{a7/a8}) as in the \textit{\textbf{top row}}, or change the cup grasped by the right arm to cups of different shapes (\textit{e.g.}, \textit{mugs with handles} in \textbf{b1/b2/b3/b4}, and \textit{ordinary cups without handles} in \textbf{b5/b6/b7/b8}) as in the \textit{\textbf{bottom row}}.}
           \label{anchorPts}
	\vspace{-15pt}
	\end{center}
\end{figure*}

\subsubsection{Representative-Point Robustness}
VLBiMan++ adopts two lightweight yet intuitive object-centric anchors: the 2D mask centroid and the foremost object-table contact point. Despite their simplicity, these anchors provide a stable interface for transferring task-relevant spatial relationships across unseen object geometries without object-specific models. We further evaluate their robustness in the \texttt{inserting} task on the platform \textbf{P2} using novel elongated objects and geometrically diverse cups and mugs, including cases with partial occlusion. As shown in Fig.~\ref{anchorPts}, the same anchoring rules remain effective across substantial shape and size variations and under dynamic perturbations. These results indicate that high-fidelity pose estimation is not essential for the considered manipulation relationships, and that simple object-centric anchors can provide an effective and scalable representation for cross-instance adaptation.

\begin{table}[t]  
	\centering
	\setlength{\tabcolsep}{1pt}
	\caption{Ablation experiments of the interpolation density $n$. To ensure reliable searching of the optimal $n$, we did not add any additional dynamic interference in each trail.}
	\vspace{-8pt}
	\begin{tabular}{c|cccc|c|cccc|c}
	\Xhline{1.2pt}
	\multirow{6}{*}{\makecell{Inter-\\polation\\Density\\$n$}} & \multicolumn{5}{c|}{\textit{\textbf{\cellcolor{blue!10}ID setting}}} & \multicolumn{5}{c}{\textit{\textbf{\cellcolor{blue!30}OOD setting}}} \\
	\cline{2-11}
	~ & \rotatebox[origin=c]{80}{\texttt{inserting}} & \rotatebox[origin=c]{80}{\texttt{unscrew}} & \rotatebox[origin=c]{80}{\texttt{pouring}} & \rotatebox[origin=c]{80}{\texttt{reorient}}
		& \makecell{~\\Average\\Success\\Rate} 
		& \rotatebox[origin=c]{80}{\texttt{inserting}} & \rotatebox[origin=c]{80}{\texttt{unscrew}} & \rotatebox[origin=c]{80}{\texttt{pouring}} & \rotatebox[origin=c]{80}{\texttt{reorient}}
 		& \makecell{~\\Average\\Success\\Rate} \\
	\Xhline{0.8pt} 
 	$n=3$ & 15/20 & 14/20 & 12/20 & 13/20 & \cellcolor{gray!15}67.5\% 
		& 13/20 & 12/20 & 11/20 & 11/20 & \cellcolor{gray!15}58.8\% \\ 
	$n=4$ & 18/20 & 15/20 & 15/20 & 16/20 & \cellcolor{gray!15}80.0\%  
		& 17/20 & 14/20 & 14/20 & 14/20 & \cellcolor{gray!15}73.8\% \\ 
	$n=5$ & 19/20 & 17/20 & 18/20 & 16/20 & \cellcolor{gray!15}87.5\%  
		& 18/20 & 16/20 & 17/20 & 15/20 & \underline{\cellcolor{gray!15}82.5\%} \\ 
	$n=6$ & 19/20 & 19/20 & 18/20 & 17/20 & \underline{\cellcolor{gray!15}91.3\%}  
		& 19/20 & 18/20 & 17/20 & 16/20 & \textbf{\cellcolor{gray!15}87.5\%} \\  
	$n=7$ & 19/20 & 18/20 & 18/20 & 18/20 & \textbf{\cellcolor{gray!15}92.5\%}  
		& 18/20 & 19/20 & 17/20 & 16/20 & \textbf{\cellcolor{gray!15}87.5\%} \\ 
	$n=8$ & 19/20 & 19/20 & 17/20 & 18/20 & \textbf{\cellcolor{gray!15}92.5\%}  
		& 19/20 & 18/20 & 16/20 & 17/20 & \textbf{\cellcolor{gray!15}87.5\%} \\  
	\Xhline{1.2pt}
	\end{tabular}
	\label{tabG}
	\vspace{-10pt}
\end{table}

\subsubsection{Interpolation Density}
We further study the interpolation density $n$ used during pre-grasping (refer Eqn.~\ref{IKrefine}). Interpolated waypoints provide a smoother and safer approach, reduce premature end-effector-object contact, and create intermediate states for responding to external or execution-induced object offset. We utilized the dual-arm humanoid platform \textbf{P2} to conduct four bimanual tasks. As shown in Tab.~\ref{tabG}, increasing $n$ generally improves grasp stability, while the benefit saturates beyond a moderate density. This indicates that VLBiMan++ does not require excessive waypoint discretization. We use $n\!=\!6$ as a practical choice that balances pre-grasping reliability and execution efficiency across tasks.

\begin{table}
	\centering
	\setlength{\tabcolsep}{1pt}
	\caption{Ablation studies of VLBiMan++. All trials were completed on six primary tasks using the platform \textbf{P1} (\textbf{ID} + \textbf{interference}).}
	\vspace{-8pt}
	\begin{tabular}{c|c|c|c|c|c}
	\Xhline{1.2pt}
	\makecell{VLMs-based\\grounding} & \makecell{state-\\aware\\adaptation} & \makecell{IK-\\refinement} & \makecell{collision\\avoidance} & \makecell{closed-loop\\re-composition} & \makecell{Avg.\\SR} \\
	\hline
	SAM+DINOv2 & Ours & \ding{51} & \ding{51} & \ding{51} & \cellcolor{gray!15}35.8\% \\  
	Ours & AnyGrasp & \ding{51} & \ding{51} & \ding{51} & \cellcolor{gray!15}31.7\% \\  
	Ours & Ours & \ding{55} & \ding{51} & \ding{51} & \cellcolor{gray!15}29.2\% \\  
	Ours & Ours & \ding{51} & \ding{55} & \ding{51} & \cellcolor{gray!15}34.2\% \\  
	Ours & Ours & \ding{51} & \ding{51} & \ding{55} & \cellcolor{gray!15}40.8\% \\  
	Ours & Ours & \ding{51} & \ding{51} & \ding{51} & \textbf{\cellcolor{gray!15}59.2\%} \\  
	\Xhline{1.2pt}
	\end{tabular}
	\vspace{-10pt}
	\label{tabF}
\end{table} 

\begin{figure}[t]
	\centering
	\includegraphics[width=0.7\columnwidth]{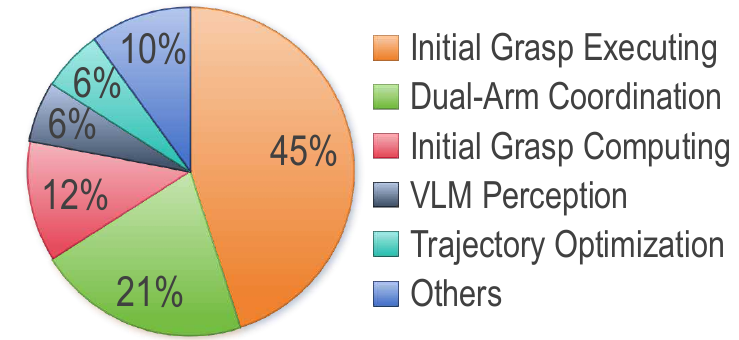}
	\vspace{-10pt}
	\caption{Error breakdown performed on VLBiMan++ failures based on the rollout results in Tab.~\ref{tabB} (covering OOD and interference part).}
	\label{errorStat}
	\vspace{-10pt}
\end{figure}

\subsubsection{Module Ablation}
We first ablate major components of VLBiMan++ by replacing or removing VLM-based grounding, state-aware adaptation, trajectory refinement (IK-refinement + collision avoidance), or closed-loop re-composition while keeping the other pipeline unchanged to examine its contribution. As shown in Tab.~\ref{tabF}, each component contributes consistently to overall performance. Replacing VLM grounding degrades semantic target localization, while replacing state-aware adaptation particularly with the popular non-semantic AnyGrasp \cite{fang2023anygrasp} (where we find the one closest to the demo grasp pose from many proposals for fair comparison) affects novel object geometries and articulated or deformable configurations. Without refined trajectory refinement, reachability and collision failures increase, whereas removing closed-loop re-composition reduces robustness to object displacement during execution. These results verify that the proposed components are complementary and collectively support VLBiMan++'s generalization and robustness.

\subsubsection{Failure Analysis}
We analyze remaining failures using four stages: {Perception} $\rightarrow$ \textit{Planning}$ \rightarrow$ \textit{Execution} $\rightarrow$ \textit{Recovery}. The major failure categories consist of five compositions, consistent with the breakdown in Fig.~\ref{errorStat}. Among them, execution-related failures remain the dominant source, reflecting the difficulty of reliable physical grasping under kinematic singularities, imperfect contact, and early collisions. Dual-arm coordination constitutes another important source of failure, whereas VLM perception and trajectory optimization account for smaller proportions, supporting the reliability of the proposed grounding and trajectory synthesis modules. In the \textit{Recovery} stage, the closed-loop mechanism can compensate for moderate object displacement and temporary perception uncertainty by re-observing and selectively re-instantiating affected skills, but it cannot yet recover from fundamentally unreachable configurations, severe grasp failures, or large unmodeled object-state changes. This analysis clarifies both the practical strengths of VLBiMan++ and the failure boundaries that remain important for future improvement.


\section{Conclusion and Limitation}
In this work, we presented VLBiMan++, an extended framework for generalizable bimanual robotic manipulation that expands the reuse boundary of a single demonstration through vision-language anchoring, state-aware skill abstraction, and closed-loop trajectory composition. By decomposing demonstrations into state-invariant and state-conditioned skills, grounding task-relevant objects through vision-language perception, and selectively adapting their geometric or object-state-dependent components, VLBiMan++ avoids repeated data collection and policy retraining while supporting increasingly diverse conditions. Extensive real-world experiments demonstrate generalization across tasks and skill compositions, unseen rigid, articulated, and deformable objects, cluttered and dynamic scenes, heterogeneous dual-arm embodiments, and prolonged closed-loop execution. These results suggest that a carefully structured one-shot demonstration can serve not merely as a trajectory to imitate, but as a reusable task prior whose functional structure can be repeatedly re-instantiated across changing physical and environmental conditions.

\textbf{Limitations}: Despite these advances, several limitations remain. \textit{\textbf{First}}, the current object-state representation is intentionally compact and task-oriented, and therefore does not fully model highly complex or dynamically changing object configurations. Although VLBiMan++ extends beyond rigid objects to articulated and deformable cases, highly deformable materials with topology changes, self-occlusion, or strong dynamics may require richer representations and more frequent perception-action updates. \textit{\textbf{Second}}, the current closed-loop mechanism primarily detects observable state deviations and selectively re-adapts the affected skills. It does not yet provide a comprehensive failure-aware reasoning and recovery system capable of diagnosing whether an action has succeeded, identifying the cause of a failure, and autonomously selecting an alternative strategy. This remains particularly important for long-horizon manipulation, where small execution errors can accumulate across multiple contacts and skill transitions. \textit{\textbf{Third}}, collision handling is still largely structured around pre-grasp compensation and validated skill trajectories. More general collision-aware planning in densely cluttered and dynamically changing environments would benefit from dedicated model-based planners or online reactive controllers that explicitly consider the full configuration space of both arms and surrounding objects. \textit{\textbf{Fourth}}, the current task decomposition still involves lightweight human refinement when automatic keypose extraction is insufficient, which limits fully autonomous scaling to large collections of demonstrations. Reliable automatic segmentation of task- and state-dependent skills remains an open problem. \textit{\textbf{Finally}}, the capability of the system is bounded by the sensing and actuation of the underlying hardware. Fixed-base arms limit workspace mobility, parallel grippers restrict dexterous in-hand manipulation, and the absence of force or tactile sensing limits interaction with force-sensitive, fragile, or contact-rich objects. Future research should therefore explore richer multimodal sensing, dexterous and force-aware end-effectors, mobile or whole-body embodiments, autonomous skill discovery, and failure-aware closed-loop recovery, with the ultimate goal of moving from one-shot skill transfer toward persistent, self-correcting, and broadly capable robotic manipulation.

\bibliographystyle{IEEEtran}
\bibliography{refs-v2}


\begin{IEEEbiography}[{\includegraphics[width=1in,height=1.25in,clip,keepaspectratio]{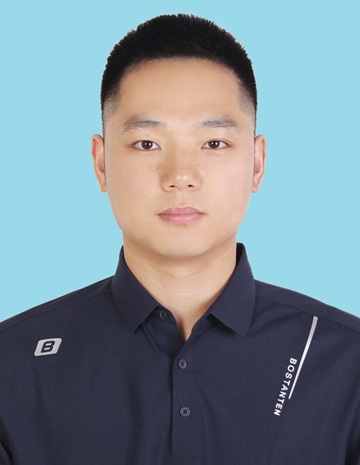}}]{Huayi Zhou}
is now an Assistant Professor at the College of Computer Science and Software Engineering, Shenzhen University. Before that, he was a postdoctoral researcher at The Chinese University of Hong Kong, Shenzhen during years 2024-2026. He received his B.S degree at Dept. of Computer Science from Hunan University in 2017, and both M.S. degree and Ph.D. degree at Dept. of Computer Science and Engineering from Shanghai Jiao Tong University in 2020 and 2024, respectively. His research interests lie in emboided robot manipulation, multi-modal perception, and computer vision (e.g., object detection, pose estimation, and monocular 3D reconstruction), combined with enduring machine learning techniques such as multi-task learning, domain adaptation, domain generalization and semi-supervised learning. 
\end{IEEEbiography}

\vspace{-10pt}

\begin{IEEEbiography}[{\includegraphics[width=1in,height=1.25in,clip,keepaspectratio]{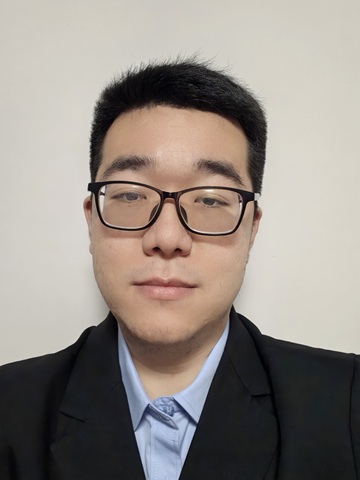}}]{Wei Gao}
is now the Chief Scientist at Dexforce Robotics. Before that, he was the Chief Scientist at Mech-Mind Robotics during years 2021-2024. He received his B.S. degree at the School of Aerospace Engineering from Tsinghua University in 2017, and his Ph.D. degree at the Department of Electrical Engineering and Computer Science from Massachusetts Institute of Technology (MIT) in 2021. His research interests lie in robotics, with sub-fields include manipulation, 3D vision, motion planning \& control, and robotic foundation models (such as Vision-Language-Action (VLA) and World-Action Model (WAM)).
\end{IEEEbiography}

\vspace{-10pt}

\begin{IEEEbiography}[{\includegraphics[width=1in,height=1.25in,clip,keepaspectratio]{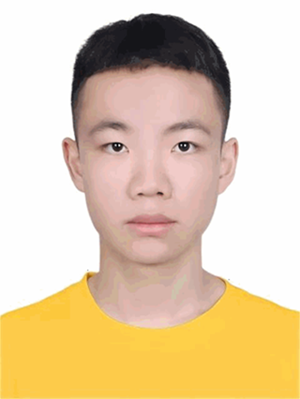}}]{Yiyang Han}
is currently an undergraduate student at the Faculty of Engineering, Department of Computing, Imperial College London. His research interests are in machine learning and artificial intelligence. His recent research focuses on real-time turn-taking in spoken dialogue systems, parameter-efficient imitation learning for human behavior modeling, and embodied robot perception and manipulation.
\end{IEEEbiography}

\vspace{-10pt}

\begin{IEEEbiography}[{\includegraphics[width=1in,height=1.25in,clip,keepaspectratio]{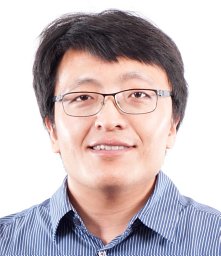}}]{Kui Jia}
is currently a Professor with School of Data Science, The Chinese University of Hong Kong, Shenzhen. He received the B.Eng. degree in marine engineering from Northwestern Polytechnical University, China, in 2001, the M.Eng. degree in electrical and computer engineering from the National University of Singapore in 2003, and the Ph.D. degree in computer science from Queen Mary University of London, London, U.K., in 2007. His research interests are in computer vision and machine learning. His recent research focuses on theoretical deep learning and its applications to 3D vision. He has been serving as an Associate Editor for IEEE Trans. on Image Processing and Trans. on Machine Learning Research.
\end{IEEEbiography}

\vspace{-10pt}

\begin{IEEEbiography}[{\includegraphics[width=1in,height=1.25in,clip,keepaspectratio]{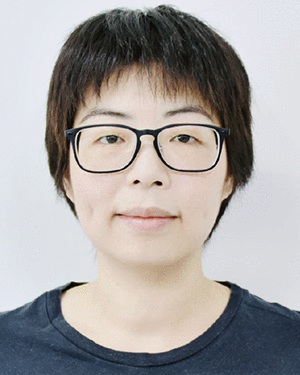}}]{Hui Huang}
is a Chair Professor at Shenzhen University, serving as the Dean of CS while also directing the Visual Computing Research Center (VCC). Her research spans computer graphics, computer vision, and visual analytics, with a particular focus on geometry. She has held editorial board positions for ACM TOG and IEEE TVCG. She is an inducted member of the ACM SIGGRAPH Academy and a CSIG Fellow. She has also been appointed Technical Papers Chair for ACM SIGGRAPH Asia 2027. Her homepage is: \href{https://vcc.tech/~huihuang}{\color{blue}https://vcc.tech/huihuang}.
\end{IEEEbiography}

\vfill

\end{document}